# Measurable Semantic 3D Gaussian Splatting with Joint Region and Depth Supervision for Interactive Building Inspection

Yi Zhang[1], Tian Tan[1], and Ray Kai Leung Su[1,*]

[1] Department of Civil Engineering, The University of Hong Kong, Hong Kong, China

**Abstract:** Image-based 3D reconstruction provides a scalable representation for building inspection and maintenance. However, photorealistic reconstruction alone does not fully provide the geometric measurability, semantic structure, and interactive information access required for engineering applications. This study proposes a measurable semantic 3D Gaussian Splatting (3DGS) framework with region and depth supervision for interactive building inspection from multi-view imagery. Mask-based region supervision suppresses irrelevant backgrounds and focuses optimization on target regions. Point-wise depth supervision guides Gaussian centers toward physical surfaces using reference geometry derived from multi-view stereo (MVS) point clouds, providing geometric anchors for measuring physical dimensions and areas. A semantic Gaussian field associates learnable class features with individual Gaussian primitives, enabling semantic rendering and component-level inspection. These capabilities are operationalized through AEC-SplatTwin, an interactive building inspection platform supporting real-time visualization, semantic analysis, quantitative measurement, and multimodal large language model (MLLM)-assisted interaction. On the synthetic UAV-inspection QuakeCity and smartphone-captured HKUPB datasets, the framework reduces Gaussian-center depth MAE by 58.1% and 86.8%, while its semantic field achieves mIoU values of 0.8529 and 0.8642. On HKUPB, the mean relative errors for height, width, and area measurements are 0.33%, 1.07%, and 0.60%. These results show that the proposed framework extends 3DGS beyond visual reconstruction into a measurable, semantically structured, and interactive engineering representation for building inspection.

# 1. Introduction

The increasing use of unmanned aerial vehicles (UAVs), handheld cameras, and mobile robots is generating large volumes of inspection images and videos in the Architecture, Engineering, and Construction (AEC) domain [1-3]. Although these platforms improve the coverage of extensive, complex, and difficult-to-access assets, their observations remain fragmented across viewpoints, making cross-view component identification, spatial defect localization, and asset-level evidence consolidation increasingly difficult [4-7]. Image-based 3D reconstruction provides a means of organizing these observations within a coherent 3D representation for inspection, assessment, and maintenance [8,9]. Classical photogrammetry pipelines combining structure-from-motion (SfM) and multi-view stereo (MVS) recover camera poses and dense point clouds that support spatial modeling and scale-calibrated measurement [10,11]. However, point clouds are often visually noisy and lack explicit semantic structure, limiting their effectiveness for intuitive inspection and high-level reasoning tasks. By combining high-quality reconstruction with real-time rendering through explicit Gaussian primitives, 3D Gaussian Splatting (3DGS) provides a promising basis for integrating inspection imagery into a spatially coherent, asset-level representation [12]. Nevertheless, an engineering-oriented representation for building inspection must support more than photorealistic visualization [13]. Depending on the application, engineers may need to isolate the target building from backgrounds and occlusions, identify components and defects, quantify physical dimensions and affected areas, compare evidence across viewpoints, and interactively access the resulting information. These are knowledge-intensive engineering tasks because they require visual observations to be linked with engineering entities, spatial geometry, and application-specific rules. The central challenge is how to transform the reconstructed 3DGS into a target-aware, measurable, semantically structured, and interactive engineering representation for building inspection.

Target-aware reconstruction presents the first challenge. Standard 3DGS applies its RGB loss to all image pixels and therefore reconstructs all visible content rather than only the target region [14]. Images collected during UAV surveys and manual or robotic inspections commonly contain neighboring buildings, sky, vegetation or foreground occlusions. These regions often have insufficient multi-view coverage and can produce unstable Gaussian primitives and reconstruction artefacts. Existing AEC studies isolate targets through image preprocessing or model post-processing, such as manually removing irrelevant Gaussians or projecting foreground-background masks into the reconstructed model [15,16]. However, these methods do not constrain the target region during Gaussian optimization, while mask projection may become unreliable under viewpoint changes, occlusions, and object overlap. An intrinsic region-aware strategy is therefore needed to reduce manual intervention and focus reconstruction on inspection-relevant regions.

Geometric measurability presents a second challenge. Each 3D Gaussian is an anisotropic volumetric primitive whose ellipsoidal spatial extent represents rendering support rather than the physical boundary of a building component. Because standard 3DGS is driven mainly by photometric losses, visually satisfactory reconstruction does not guarantee that Gaussian primitives align with physical surfaces. Gaussian primitives may remain distributed around the actual building surfaces, while their ellipsoidal extents cannot be directly interpreted as component dimensions. Existing depth-supervised 3DGS methods generally constrain an alpha-composited depth map using scale-aligned or normalized monocular depth priors [17-19]. However, rendered depths are jointly determined by Gaussian position, scale, opacity, and transmittance. Reducing rendered-depth error therefore does not necessarily establish explicit, surface-aligned geometry suitable for measuring physical dimensions and areas. Yu et al. showed that reliable geometric information for direct measurement had to be obtained from LiDAR rather than 3DGS itself in AEC applications [20]. The unresolved problem is how measurable geometry can be defined within the Gaussian representation and aligned with physical surfaces.

Semantic information introduces a third representational requirement. Structural components and defects, such as beam, window, cracking, and spalling, constitute the fundamental units for building inspection, condition assessment, and maintenance planning [21]. Associating these entities with a 3D representation also provides an interface to downstream information environments, including Building Information Modeling (BIM), Industry Foundation Classes (IFC), and multimodal large language model (MLLM)-assisted reasoning systems [22-25]. Computer vision studies have embedded semantic or language features into Gaussian primitives for open-vocabulary scene understanding and querying [26-30]. In AEC applications, semantic masks and damage labels have been used for target reconstruction and damage visualization [15,16], while Heritage-3DGS and VS-BIM further advanced visual representations toward domain-specific semantics and MLLM-assisted information access [6,14]. For an engineering representation intended for building inspection, semantic information should describe what individual Gaussian primitives represent and connect this information with measurable geometry, rather than remaining an image-level visualization result.

Beyond these representation-level questions, the engineering value of a semantic 3DGS representation depends on whether its visual, geometric, and semantic capabilities can be operationalized within an interactive platform. Existing AEC-oriented 3DGS systems are largely limited to preprocessing and basic visualization of reconstructed 3DGS models, without support for key functionalities, such as integrated quantitative measurement, semantic inspection, and natural-language access to scene information. An operational platform is therefore required to translate the proposed representation into task-oriented functions for building inspection and engineering analysis.

In summary, previous studies advanced 3DGS from visual appearance to semantic representation, while engineering geometry had yet to become directly measurable within 3DGS. Our study extends this established line to an integrated chain of appearance-semantics-geometry-platform. The scientific contribution lies in establishing a unified, physically measurable, and operational engineering representation for practical building inspection. Accordingly, four research questions are formulated:

1. How can 3DGS be effectively and efficiently constrained to reconstruct only target regions?
2. How can measurable geometry be established within 3DGS and aligned with physical surfaces?

3. How can 3DGS encode component- and defect-level semantic information?
4. How can the resulting visual, geometric, and semantic information be operationalized within an interactive platform for building inspection?

To address these questions, this study proposes a measurable semantic 3DGS framework for interactive building inspection. The principal technical innovations are as follows:

1. A mask-based region-supervision strategy defines a shared valid domain for RGB and depth optimization, restricting 3DGS reconstruction to target regions and suppressing irrelevant backgrounds and foreground occlusions.
2. A point-wise depth-supervision method defines Gaussian centers as explicit geometric anchors and constrains their camera-space depths using MVS-derived reference geometry, guiding them toward physical surfaces and providing a geometric basis for measuring physical dimensions and areas.
3. A semantic Gaussian field binds learnable semantic features to individual Gaussian primitives, enabling semantic rendering and component-level inspection.
4. AEC-SplatTwin, an interactive building inspection platform developed in this study, operationalizes the proposed representation through real-time visualization, semantic analysis, distance and area measurement, and MLLM-assisted interaction.

The proposed framework is evaluated on the synthetic QuakeCity dataset and the smartphone-captured HKUPB dataset. At the representation level, target-region RGB reconstruction, MVS-referenced Gaussian-center alignment, and semantic accuracy are evaluated. At the application level, physical measurement accuracy and the interactive functions of AEC-SplatTwin are demonstrated.

This paper is organized as follows. Section 2 reviews related work on 3D Gaussian Splatting, semantic Gaussian representations, and real-time interactive systems for AEC applications. Section 3 introduces the proposed framework, including point-wise depth supervision, mask-based region supervision, semantic Gaussian field construction, and AEC-SplatTwin. Section 4 describes the datasets, evaluation metrics, and experimental setup. Section 5 presents the results and discussion on RGB reconstruction, geometric consistency, and semantic rendering. Section 6 introduces the real-time visualization, interactive measurement, and MLLM-assisted interaction functions of AEC-SplatTwin. Section 7 discusses limitations and future work, and Section 8 concludes the paper.

## 2. Related Work

### 2.1 3D Gaussian Splatting

Image-based sensing has become an important source of engineering information for buildings and infrastructure. UAVs support rapid inspection of large or difficult-to-access surfaces, while handheld cameras and mobile robots provide flexible data collection in local, indoor, or constrained environments. SfM and MVS can integrate these observations by recovering camera poses and generating sparse or dense point clouds [11,31]. Neural rendering methods have significantly improved the realism and efficiency of image-based reconstruction [32]. Kim and Cha combined a modified Nerfacto model with STRNet-based crack segmentation and test-time augmentation to reconstruct a bridge model and map cracks [33]. Alongside NeRF-based approaches, 3DGS offers an explicit scene representation with real-time rendering capability and high-fidelity texture preservation [12].

Despite these advantages, standard 3DGS is optimized for visual appearance rather than geometric accuracy. The ellipsoidal extent of each Gaussian represents rendering support rather than a physical surface boundary, while the optimized Gaussian distribution is not guaranteed to align with the underlying physical surfaces [34]. Depth-Regularized 3DGS [17] aligns monocular depth estimates with sparse SfM points, CDGS [18] further introduces confidence-aware weighting to reduce the influence of unreliable monocular depth estimates, and DNGaussian [19] employs hard and soft depth regularization with global-local normalization to improve Gaussian positioning under sparse-view inputs. These methods improve novel-view synthesis and geometric consistency, but their objectives are generally defined on alpha-composited rendered depth. They do not explicitly establish and validate Gaussian-center geometry for physical measurement in AEC applications.

Recent AEC studies have applied 3DGS to heritage digitization, building reconstruction, energy modeling, and structural defect inspection. Duan et al. combined Segment Anything Model (SAM)-based segmentation with 3DGS to reconstruct concrete structures from monocular video, enabling defect-aware 3D visualization [15], while Wang et al. [16] incorporated 2D damage information into 3DGS-based reconstruction to support multi-scale damage visualization and model updating. Zhang et al. [35] proposed DGE-GS to improve damage fidelity and geometric consistency in post-earthquake building reconstruction. Yu et al. [20] compared 3DGS and LiDAR within heritage digitization workflows and demonstrated its advantages in photorealistic visualization and interactive rendering, while noting the greater geometric reliability of LiDAR point clouds. Liang et al. [14] further extended 3DGS toward semantic and interactive applications by integrating multimodal large language models, enabling language querying and scene exploration. More recent studies have extended these applications to building surface meshing and building energy modeling through geometric feature extraction [36,37]. These works demonstrate the engineering potential of 3DGS, but most emphasize reconstruction, visualization, or downstream model generation. Target-region constraints during Gaussian optimization and surface-aligned Gaussian geometry validated for physical measurement remain insufficiently investigated.

## 2.2 Semantic Gaussian representations

Semantic reconstruction organizes visual observations according to engineering entities such as components, equipment, and defects [22]. Image segmentation methods and foundation models [38-40] such as SAM provide dense semantic cues in individual views, whereas point-cloud segmentation methods [41-45], including PointNet and PointNet++, directly classify 3D points. However, independently segmenting images or point clouds separates semantic information from the scene representation used for photorealistic rendering and interaction [46].

Recent computer vision studies have therefore begun to explore semantic- or language-embedded neural scene representations [26,47]. LangSplat [28] and LangSplatV2 [29] embed language-aligned features in Gaussian representations for open-vocabulary querying. Dr. Splat [27] further advances this direction by directly registering language-aligned embeddings to Gaussian primitives, while ReferSplat [30] extends Gaussian representations to spatially aware referring segmentation and language-guided reasoning. These methods demonstrate that Gaussian primitives can simultaneously carry appearance and semantic information, although they primarily target general-purpose scene understanding.

AEC studies have also begun to incorporate semantic information into Gaussian scenes. Damage masks have been projected into reconstructed structures for defect visualization [15,16], while Heritage-3DGS [14] constructs a domain-specific language-embedded Gaussian field for semantic retrieval of historical-building components. Therefore, the remaining research gap is that existing semantic studies do not fully integrate semantic features with target-focused reconstruction, surface-aligned measurable geometry, and component-level engineering operations. The present study addresses this gap by associating semantic classes with the same Gaussian primitives used for geometric measurement and interactive visualization.

### 2.3 Real-time interactive system for AEC application

Interactive systems transform digital building representations from static models into operational platforms. Recent research has emphasized that engineering digital twins should move beyond visual representations and support monitoring, analysis, and application-oriented workflows. Existing platforms commonly use BIM, point clouds, or parametric models as their primary representation. For example, Lv et al. [48] integrated automatically generated BIM models into a web-based digital twin platform for interactive visualization and attribute querying of engineered slopes. Such systems demonstrate the value of real-time access to asset information but remain dependent on predefined model structures.

Natural-language interaction has recently been introduced to improve access to complex spatial and engineering information [49]. Zhu et al. [50] integrated large language models with GIS and knowledge graphs to support natural-language-driven spatial analysis. Graph-based digital twin and openBIM studies have further applied LLM agents to spatial querying, carbon assessment, and engineering information retrieval [51,52]. Lu et al. [53] developed an LLM-driven multi-agent framework to reduce the cognitive and operational burden associated with conventional digital twin interfaces. These studies demonstrate the importance of linking visual data with semantic and spatial information for knowledge-intensive engineering tasks. However, they generally operate on BIM, GIS, point-cloud, or graph representations rather than a photorealistic and measurable Gaussian scene.

Within 3DGS-based AEC applications, Heritage-3DGS [14] provides chatbot-assisted semantic querying of historical buildings, while Gaussian Splatting-enabled structural digital twins[16] support damage visualization and model updating. Nevertheless, real-time visualization, semantic inspection, quantitative measurement, and MLLM-assisted interaction have generally been developed as separate capabilities. Their integration within a single 3DGS-based environment for building inspection, particularly one supported by measurable Gaussian geometry, has received limited attention. This motivates AEC-SplatTwin, which connects the visual, geometric, and semantic properties of the proposed representation with interactive engineering functions. Table 1 compares the inputs, engineering scenarios, and principal capabilities of related AEC studies.

## 3. Method

As illustrated in Fig. 1, standard 3DGS serves as the photometric reconstruction backbone of the proposed framework. Point-wise depth supervision is first introduced to constrain the camera-space depths of Gaussian centers using MVS-derived depth maps, guiding them toward physical surfaces and establishing a geometric basis for subsequent measurement. Because mask-based region supervision operates jointly on the photometric and geometric objectives, it is subsequently applied to both the RGB and depth losses to define a shared valid domain, thereby restricting optimization to the target structure and suppressing irrelevant backgrounds and foreground occlusions. Based on the resulting target-focused and measurable 3DGS representation, learnable semantic features are attached to individual Gaussian primitives and optimized through semantic rendering with segmentation-label supervision. Finally, the measurable semantic 3DGS representation is deployed in AEC-SplatTwin, which provides real-time multimodal visualization, semantic inspection, interactive distance and area measurement, and MLLM-assisted interaction.

Table 1. Comparison of 3DGS-related studies for AEC applications.

| Study | Input | Engineering scenario | Target reconstruction | Physical measurement | Semantic representation | Interactive platform | MLLM |
|---|---|---|---|---|---|---|---|
| Wang et al. [6] | Video; BIM | Construction safety inspection | – | – | ✓ | – | ✓ |
| Liang et al. [14] | Image | Heritage digitization | – | – | ✓ | – | ✓ |
| Duan et al. [15] | Image | Building defect inspection | – | – | – | – | – |
| Wang et al. [16] | Image | Building defect inspection | – | – | ✓ | – | – |
| Yu et al. [20] | Image; LiDAR | Heritage digitization | – | – | ✓ | ✓ | – |
| Kim et al. [33] | Image | Bridge reconstruction and crack mapping | – | – | ✓ | ✓ | – |
| Zhang et al. [35] | Image | Post-earthquake building reconstruction | – | – | ✓ | – | – |
| Wang et al. [36] | Image | Building reconstruction | ✓ | – | – | – | – |
| Liang et al. [37] | Image | Building energy modeling | ✓ | – | – | – | – |
| Lv et al. [48] | LiDAR | Slope digital twin | – | – | ✓ | ✓ | – |
| **Proposed Framework** | **Image** | **Building inspection** | ✓ | ✓ | ✓ | ✓ | ✓ |

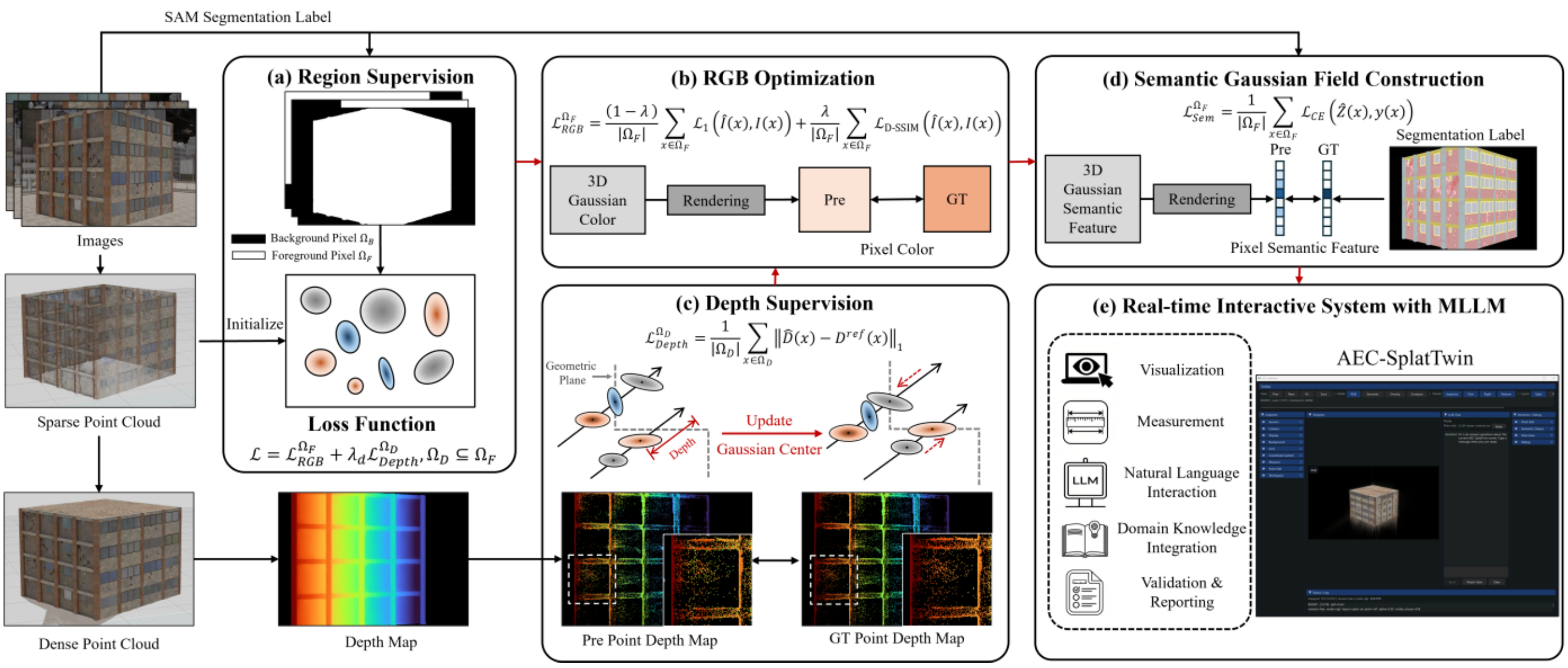

Fig. 1. Proposed framework.

### 3.1 Preliminaries of 3D Gaussian Splatting

3DGS represents a 3D scene as a set of anisotropic Gaussian primitives. Each Gaussian primitive is parameterized by its center $\mu \in R^3$, scale $s \in R^3$, rotation $q \in R^4$, opacity $\alpha$, and color feature $c$. The spatial influence of the $i$-th Gaussian is defined as:

$$G_i(x) = \exp\left(-\frac{1}{2}(x-\mu_i)^T \Sigma_i^{-1}(x-\mu_i)\right) \tag{1}$$

where the covariance matrix $\Sigma_i$ is constructed from the scale $s_i$ and rotation $q_i$.

During rendering, each 3D Gaussian primitive is projected onto the image plane, and its contribution to a pixel is determined by the projected opacity:

$$\alpha_i'(x) = \alpha_i \cdot G_i^{2D}(x) \tag{2}$$

After the projected Gaussians are ordered from front to back along the viewing direction, the rendered color at pixel $x$ is computed through $\alpha$-blending:

$$C(x) = \sum_i c_i \alpha_i'(x) \prod_{j<i} \left(1-\alpha_j'(x)\right) \tag{3}$$

where $c_i$ denotes the view-dependent color of the $i$-th Gaussian, and $\alpha_i'(x)$ denotes its projected opacity at pixel $x$.

Following the original 3DGS formulation, the photometric reconstruction loss combines an $L_1$ term with a D-SSIM term:

$$\mathcal{L}_{RGB} = (1-\lambda)\mathcal{L}_1(\hat{I}, I) + \lambda\mathcal{L}_{\text{D-SSIM}}(\hat{I}, I) \tag{4}$$

where $\hat{I}$ and $I$ denote the rendered and ground-truth images, respectively, and $\lambda$ controls the balance between the two loss terms.

3.2 Point-wise depth supervision

Instead of supervising an alpha-composited rendered depth map, the proposed method directly constrains the camera-space depths of Gaussian centers. For each pixel $x$ with an available MVS-derived reference depth $D^{ref}(x)$, Gaussian centers within the corresponding camera-ray neighborhood are identified, and their camera-space Z-depths are sorted from front to back. Candidates satisfying $d_k > D^{ref}(x) + \delta$ are discarded as geometrically inconsistent, where $\delta$ is the depth-consistency threshold. Up to $K$ frontmost candidates are then retained, and $K_x \leq K$ denotes the number of candidates remaining at pixel $x$. Their depths are aggregated to obtain the point-wise depth estimate $\widehat{D}(x)$:

$$\widehat{D}(x) = \sum_{k=1}^{K_x} \frac{\tau(1-\tau)^{k-1}}{\sum_{m=1}^{K_x} \tau(1-\tau)^{m-1}} d_k \tag{5}$$

where $d_k$ denotes the camera-space z-depth of the $k$-th retained Gaussian center, and $\tau$ is a front-to-back decay coefficient that assigns larger weights to candidates closer to the camera. The normalization ensures that the weights of the retained candidates sum to one.

The per-pixel depth discrepancy is defined as the $L_1$ distance between the aggregated point-wise depth and the MVS-derived reference depth:

$$\mathcal{L}_{Depth}(x) = \left\| \widehat{D}(x) - D^{ref}(x) \right\|_1 \tag{6}$$

where $\widehat{D}(x)$ denotes the point-wise depth aggregated from retained Gaussian centers, and $D^{ref}(x)$ denotes the reference depth derived from the MVS point cloud.

Because $\widehat{D}(x)$ is computed directly from Gaussian-center depths and is independent of Gaussian opacity and covariance, the resulting depth loss acts directly on the center positions rather than being reduced through opacity adjustment or primitive-shape deformation. The Gaussian centers are therefore guided toward the MVS-reconstructed reference surfaces, providing more reliable geometric anchors for subsequent measurement.

3.3 Mask-based region supervision

To reduce background interference and focus the optimization on the target object, a mask-guided supervision strategy is adopted. Let $M(x) \in \{0,1\}$ denote the binary foreground mask and define the foreground pixel set as $\Omega_F = \{x|M(x) = 1\}$.

The RGB reconstruction loss defined in Section 3.1 is applied only within the foreground region. Specifically, both the $L_1$ and D-SSIM terms are restricted to pixels $x \in \Omega_F$, yielding:

$$\mathcal{L}_{RGB}^{\Omega_F} = \frac{(1-\lambda)}{|\Omega_F|} \sum_{x \in \Omega_F} \mathcal{L}_1\left(\hat{I}(x), I(x)\right) + \frac{\lambda}{|\Omega_F|} \sum_{x \in \Omega_F} \mathcal{L}_{\text{D-SSIM}}\left(\hat{I}(x), I(x)\right) \tag{7}$$

In addition to photometric supervision, the foreground mask also serves as a spatial constraint for depth supervision. The depth loss is first restricted to the foreground region and is then further filtered according to geometric consistency along camera rays, resulting in a valid supervised pixel set $\Omega_D = \left\{x \in \Omega_F \middle| D^{ref}(x)\ is\ available\ and\ K_x > 0\right\}$. The depth loss is therefore computed as:

$$\mathcal{L}_{Depth}^{\Omega_D} = \frac{1}{|\Omega_D|} \sum_{x \in \Omega_D} \mathcal{L}_{Depth}(x) \tag{8}$$

Unlike pipelines that explicitly remove backgrounds before reconstruction, the proposed formulation retains the original input images and restricts supervision through foreground masks. Excluding background pixels from the photometric and depth objectives reduces the optimization and densification pressure in irrelevant regions. Combined with the standard opacity-based pruning mechanism, this encourages a target-focused Gaussian distribution without requiring manual post-processing removal. As a result, the optimized representation remains focused on the target structure and becomes less affected by background artifacts. The final training loss is formulated as:

$$\mathcal{L} = \mathcal{L}_{RGB}^{\Omega_F} + \lambda_d \mathcal{L}_{Depth}^{\Omega_D} \tag{9}$$

where $\lambda_d$ controls the relative contribution of depth supervision.

This unified mask strategy ensures that both photometric and geometric supervision are restricted to target-relevant regions, while depth supervision is further applied only to geometrically reliable pixels. Consequently, optimization becomes more stable, target-aware, and suitable for constructing a measurable 3DGS representation for building inspection applications.

### 3.4 Semantic Gaussian field construction

Based on the optimized Gaussian representation, a semantic field is constructed by associating semantic features with individual Gaussian primitives. This formulation enables geometry, appearance, and semantics to be jointly represented within a unified 3DGS representation. Specifically, each Gaussian primitive is augmented with an additional semantic feature vector $f_i \in R^C$, where $C$ denotes the semantic feature dimension. The parameters of the $i$-th Gaussian are extended as: $G_i = \{\mu_i, \Sigma_i, \alpha_i, c_i, f_i\}$. During semantic rendering, the semantic features are accumulated in a manner analogous to color compositing. For a given pixel $x$, the rendered semantic feature is computed as:

$$\hat{Z}(x) = \sum_{i} f_i \alpha_i'(x) \prod_{j<i} \left(1 - \alpha_j'(x)\right) \tag{10}$$

where $f_i$ and $\alpha_i'(x)$ denote the semantic feature vector and projected opacity of the $i$-th Gaussian, respectively.

To supervise semantic field learning, 2D segmentation labels are introduced as pixel-level supervision. The rendered semantic feature $\hat{Z}(x)$ is further converted into a discrete semantic prediction by applying an argmax operation along the channel dimension, assigning each pixel to the category with the highest response. The semantic loss is defined as the cross-entropy loss between the rendered semantic feature and the ground-truth semantic label over the foreground region:

$$\mathcal{L}_{Sem}^{\Omega_F} = \frac{1}{|\Omega_F|} \sum_{x \in \Omega_F} \mathcal{L}_{CE}\left(\hat{Z}(x), y(x)\right) \tag{11}$$

where $\hat{Z}(x)$ denotes the rendered semantic feature at pixel $x$, and $y(x)$ denotes the ground-truth semantic class label at foreground pixel $x$.

This formulation embeds semantic information directly into Gaussian primitives, enabling semantic-aware rendering and spatial querying. Since semantic features are attached to individual Gaussians, the resulting representation supports efficient mapping between 2D observations and 3D structures.

### 3.5 AEC-SplatTwin: System architecture and engineering workflow

To operationalize the proposed measurable semantic 3DGS representation, this study develops AEC-SplatTwin, an interactive platform for building inspection. The system connects the optimized Gaussian representation with engineering tasks that cannot be supported by photorealistic rendering alone, including component and defect inspection, geometric measurement, and task-oriented information access. Rather than importing the reconstructed model into an independent visualization engine, AEC-SplatTwin directly loads the optimized Gaussian attributes, semantic features, camera parameters, and scene-scale calibration information, while reusing the CUDA-based rasterization pipeline employed during model training and evaluation. This implementation maintains consistency between offline optimization and online interaction and supports interactive rendering of RGB appearance, semantic information, and Gaussian-center geometry.

Functionally, AEC-SplatTwin comprises four modules: multimodal visualization, semantic inspection, geometric measurement, and MLLM-assisted interaction. All four modules operate on the same set of Gaussian primitives, allowing visual appearance, geometric anchors, and semantic classes to remain spatially associated throughout the engineering workflow. This shared representation enables users to move from visual observation to component identification, quantitative measurement, and task-oriented interpretation without transferring information between disconnected models or software environments.

First, AEC-SplatTwin supports multimodal visualization of the optimized semantic 3DGS model, including RGB rendering, semantic rendering, semantic overlay, and Gaussian point visualization. RGB rendering provides photorealistic inspection of the reconstructed scene, while semantic rendering visualizes the learned semantic field by converting rendered per-pixel semantic features into discrete category labels. The semantic overlay mode combines semantic predictions with RGB appearance, allowing users to examine the spatial consistency between visual reconstruction and semantic understanding. In addition, the Gaussian point mode projects Gaussian centers into the current view and can visualize them according to their semantic categories.

Second, AEC-SplatTwin enables semantic analysis at both pixel and point levels. Users can selectively render, highlight, or inspect specific semantic categories in either image space or Gaussian point space, supporting targeted inspection of building components and scene elements. Based on category-specific Gaussian centers, intraclass point clustering can be further performed with geometric constraints, such as dominant plane restrictions, to approximate instance-level segmentation. The resulting point instances can support downstream operations such as bounding box generation, component-level measurement, and spatial relationship analysis.

Third, the geometric measurement module provides point-based distance measurement and pixel-based area estimation. In point-based measurement, users select Gaussian centers as geometric anchors and calculate their spatial distance. In pixel-based measurement, the system combines the rendered depth, camera parameters, and either a semantic category or an interactively selected region of interest to estimate the corresponding physical area. A scale-calibration factor derived from known-size reference objects or predefined measurement tags converts normalized model coordinates into real-world units. These complementary measurement modes translate the optimized Gaussian geometry into quantitative information for engineering tasks such as component dimension checking and defect-extent assessment.

Fourth, the MLLM-assisted interaction module provides a natural-language interface to the visualization, semantic inspection, and measurement functions. It interprets user instructions, invokes the corresponding system operations, and integrates the current rendered view, semantic distributions, measurement outputs, and predefined engineering rules to generate task-oriented responses. Importantly, the MLLM does not directly generate geometric measurements; quantitative results are produced by the deterministic measurement module and are subsequently provided to the MLLM for explanation and rule-based interpretation. This separation improves the traceability of engineering outputs while retaining the accessibility of natural-language interaction.

Overall, AEC-SplatTwin converts the proposed target-focused, measurable, and semantically structured Gaussian representation into an operational environment for building inspection. The system links visual evidence, semantic entities, geometric measurements, and natural-language access within a unified workflow, thereby addressing the fourth research question concerning the engineering operationalization of 3DGS for building inspection. Section 6 demonstrates these capabilities through multimodal inspection, interactive measurement, and MLLM-assisted engineering tasks.

# 4. Experimental design

## 4.1 Datasets and engineering inspection scenarios

Building inspection imagery is commonly acquired under two representative engineering scenarios. The first involves UAV-based inspection of large-scale or difficult-to-access building façades, where images are captured from multiple stand-off viewpoints to provide broad exterior coverage. The second involves close-range inspection using mobile robots or manually operated cameras, which enables flexible observation of local building components but may involve irregular camera trajectories, foreground occlusions, and incomplete viewpoint coverage. To evaluate the proposed framework under these complementary scenarios, the QuakeCity dataset was used to represent UAV-based inspection, whereas the HKU Press

Building (HKUPB) dataset, collected from handheld smartphone videos, was used to represent close-range manual inspection. Together, the two datasets support evaluation under both controlled synthetic conditions and practical real-world acquisition conditions.

The QuakeCity dataset represents the UAV-based inspection scenario under controlled synthetic conditions. It was generated using the physics-based graphics models developed by Hoskere [54,55]. It contains 4,809 high-resolution images with a resolution of 1920 × 1080 rendered from simulated UAV inspections of earthquake-damaged urban buildings under diverse viewpoints, lighting conditions, and damage levels. The dataset provides pixel-level annotations for structural components and damage types, including cracking, spalling, and exposed rebar. In this study, 319 RGB images of a four-story building under daytime illumination were selected from the QuakeCity dataset, as illustrated in Fig. 2. To adapt the original annotations to the semantic field learning objective, the component labels, cracking labels, and spalling labels were merged into a unified segmentation image. The selected annotations were converted into two forms: binary target masks for region and depth supervision, and multi-class segmentation maps for semantic-field learning. The RGB images were used for 3DGS reconstruction, while the corresponding masks and semantic labels supported target-region optimization and semantic Gaussian field construction, respectively.

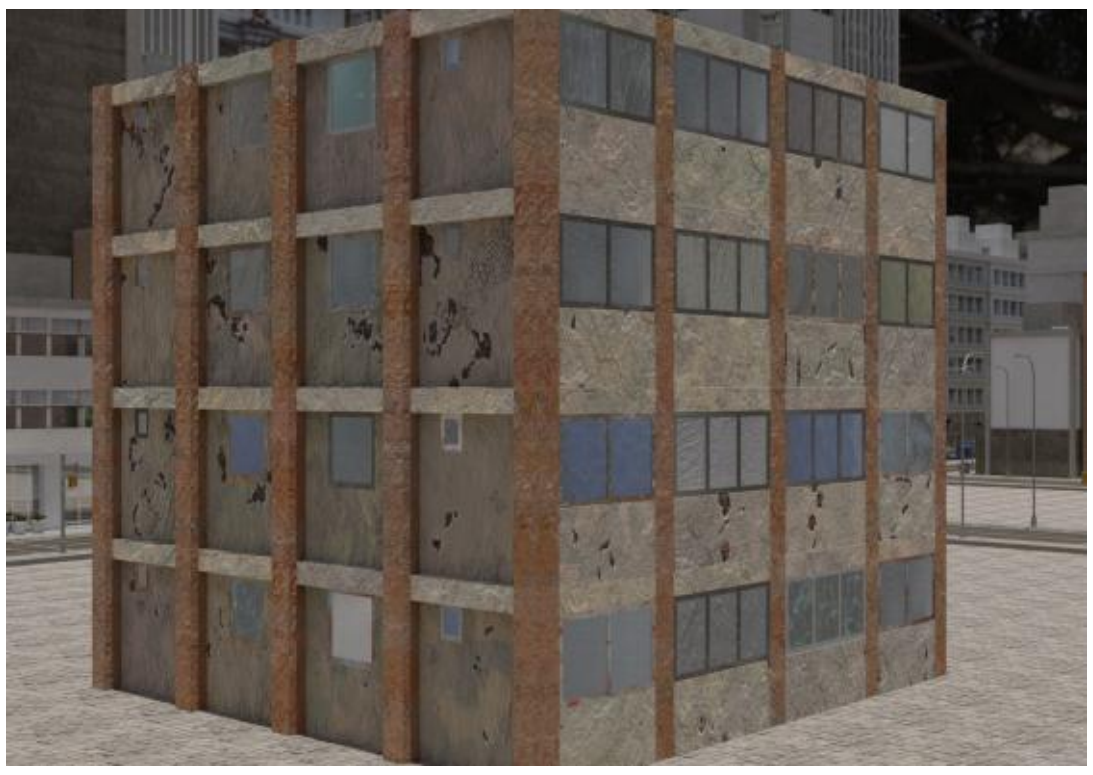

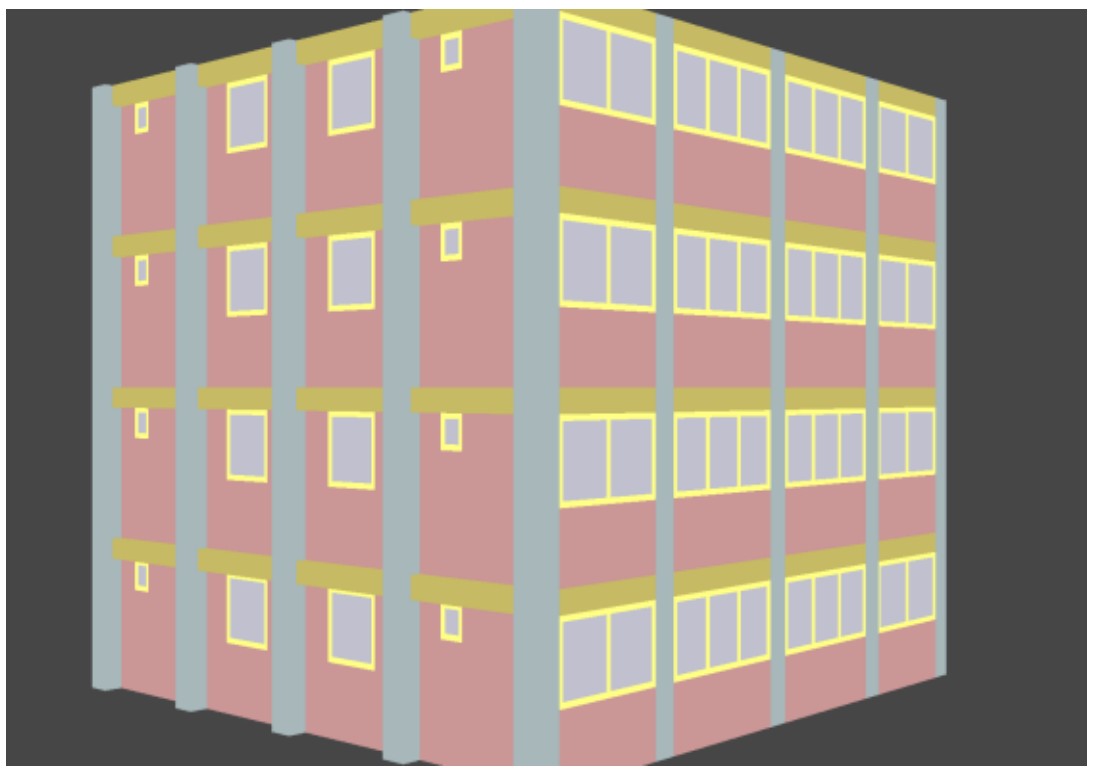

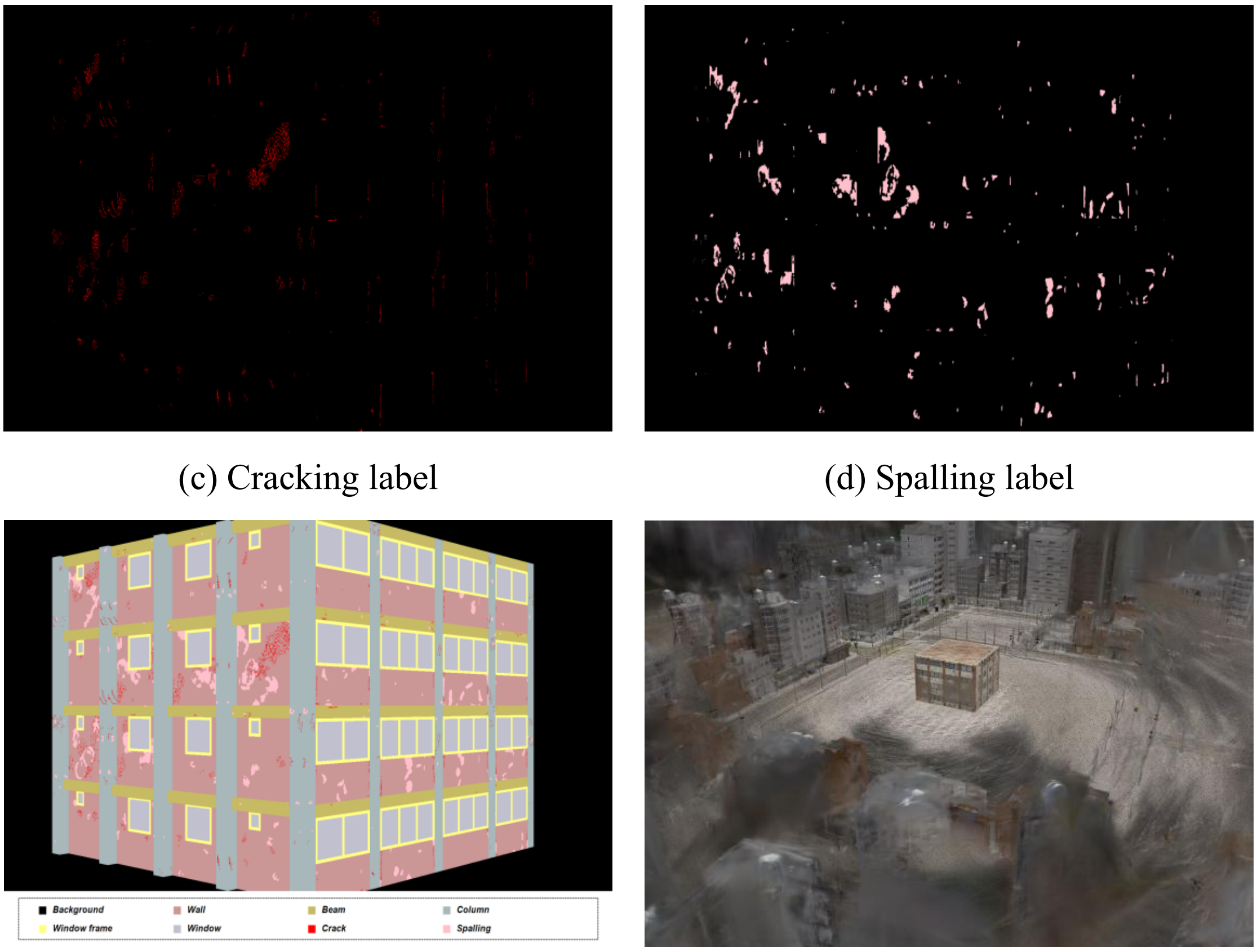

(a) RGB image (b) Component label

(c) Cracking label (d) Spalling label

(e) Unified segmentation label (f) QuakeCity 3DGS model

Fig. 2. QuakeCity dataset.

The HKUPB dataset represents close-range manual inspection under uncontrolled real-world conditions. It was collected at the HKU Press Building using a handheld smartphone, as illustrated in Fig. 3, under natural outdoor conditions with varying viewpoints. The input data were obtained by uniformly extracting frames from several handheld videos at one frame per second, resulting in 257 images with a resolution of 1920 × 1080. To evaluate the proposed framework in a real-world building scenario, semantic annotations were efficiently generated using a SAM 3 annotation tool.[40] Given the complexity of real building scenes, including occlusions, vegetation, shadows, and irregular façade details, exhaustive pixel-level annotation of all visible objects was not performed. Instead, representative categories required for system validation were annotated, including background, wall, window, column, nameplate, and round window. These selected annotations were used to support semantic field construction and to validate semantic rendering and measurement in real-world scenarios. It should be noted that the 3DGS model shown in Fig. 3(c) was generated using the original 3DGS pipeline without the proposed region/depth supervision. For clearer visualization of the complete building façade, foreground occluding structures (columns from adjacent buildings) were removed during post-processing.

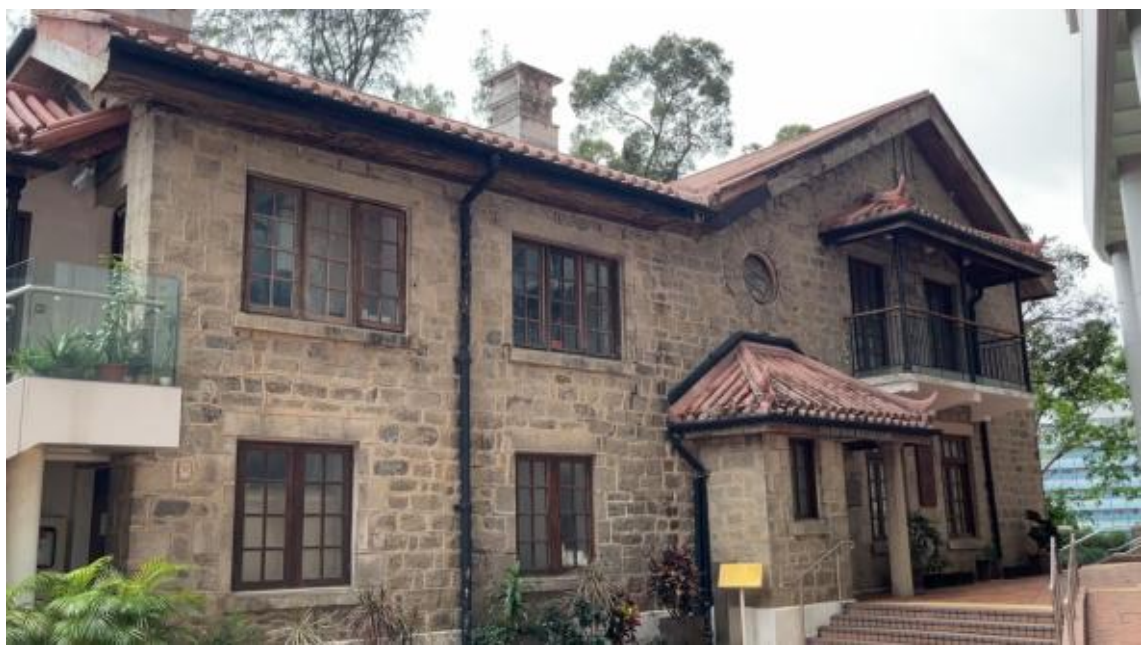

(a) RGB image

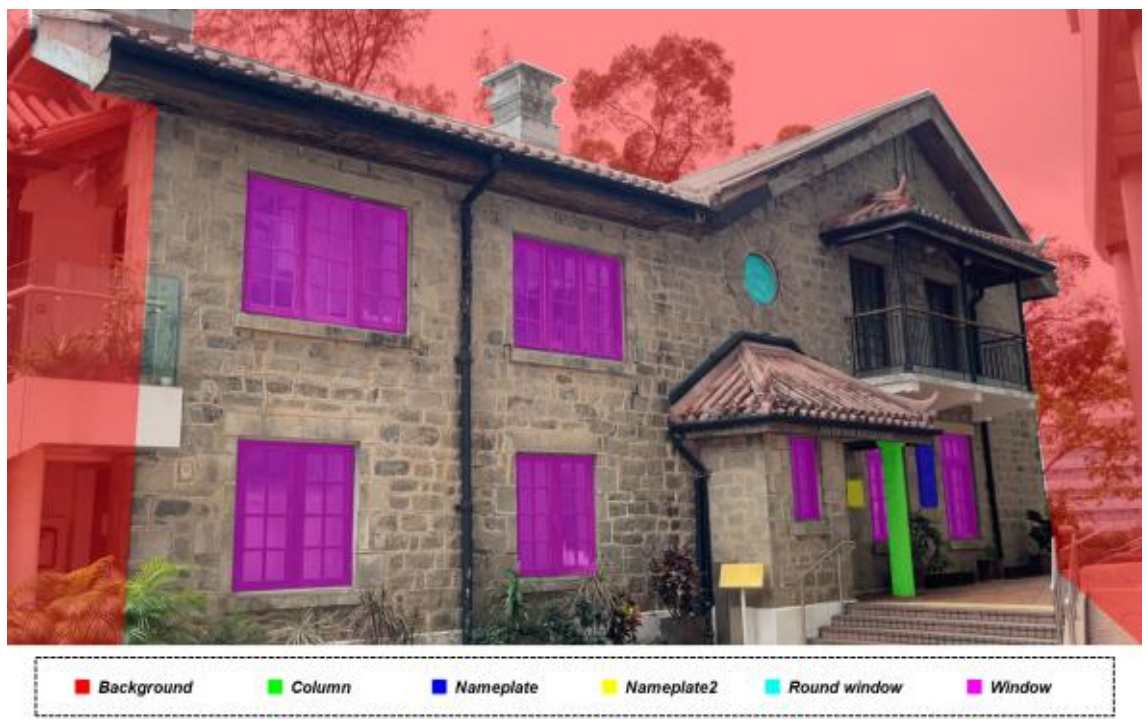

(b) Segmentation label

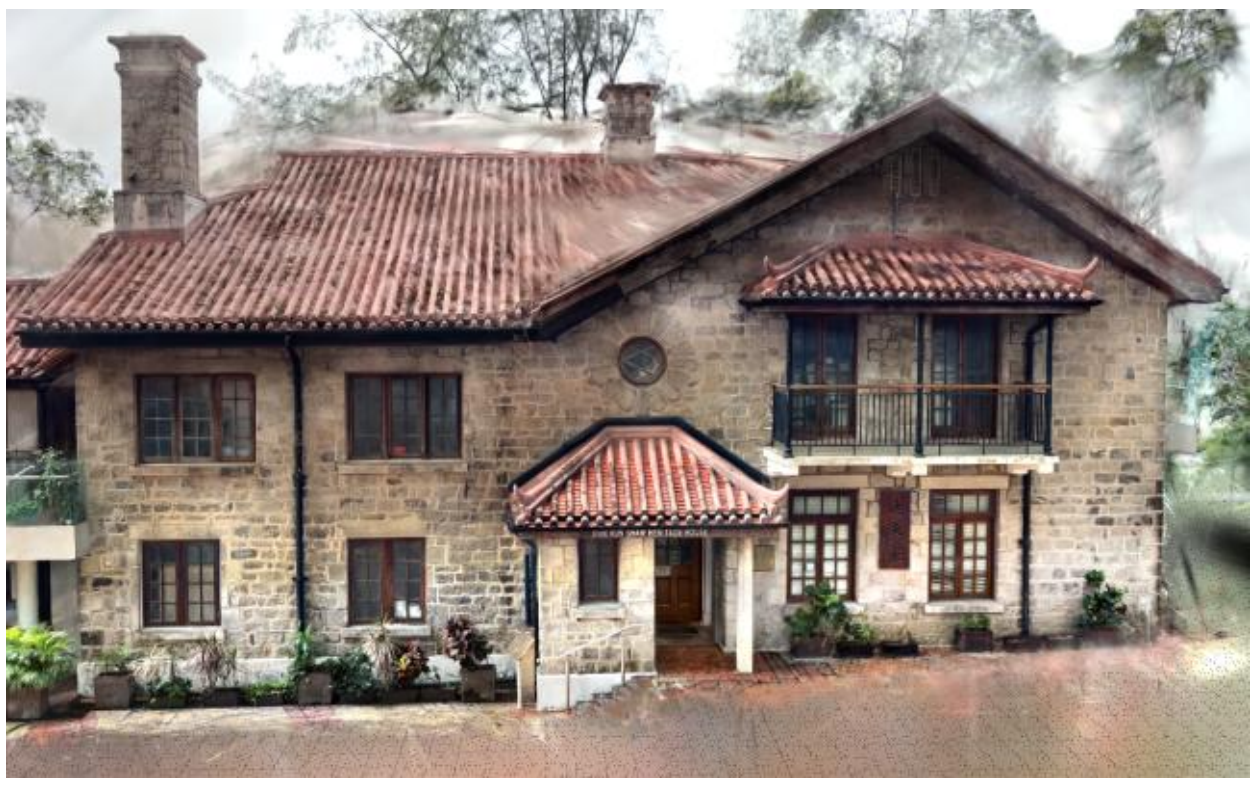

(c) HKUPB 3DGS model after removing foreground obstructions

Fig. 3. HKUPB dataset.

## 4.2 Evaluation metrics

### *4.2.1 RGB reconstruction quality*

In this study, the visual quality of RGB reconstruction is quantitatively evaluated using three widely adopted image quality metrics: Peak Signal-to-Noise Ratio (PSNR), Structural Similarity Index Measure (SSIM), and Learned Perceptual Image Patch Similarity (LPIPS).

These metrics provide complementary assessments from pixel-level accuracy, structural consistency, and perceptual similarity, respectively.

PSNR measures the pixel-wise reconstruction error between the rendered image $\hat{I}$ and the ground-truth image $I$. It is defined as:

$$\mathrm{PSNR} = 10 \cdot \log_{10}\left(\frac{\mathrm{MAX}^2}{\mathrm{MSE}}\right), \mathrm{MSE} = \frac{1}{N}\sum_{i=1}^{N}\left(\hat{I}_i - I_i\right)^2 \tag{12}$$

where MAX denotes the maximum possible pixel value, and $N$ is the total number of pixels. A higher PSNR value indicates lower pixel-wise reconstruction error and better rendering quality.

SSIM evaluates structural similarity by comparing local luminance, contrast, and structural information between $\hat{I}$ and $I$. It is formulated as:

$$\mathrm{SSIM}\left(\hat{I}, I\right) = \frac{(2\mu_{\hat{I}}\mu_I + C_1)(2\sigma_{\hat{I}I} + C_2)}{\left(\mu_{\hat{I}}^2 + \mu_I^2 + C_1\right)\left(\sigma_{\hat{I}}^2 + \sigma_I^2 + C_2\right)} \tag{13}$$

where $\mu$, $\sigma^2$, and $\sigma_{I\hat{I}}$ denote the mean, variance, and covariance of the images, respectively, and $C_1$ and $C_2$ are small constants for numerical stability. SSIM values range from 0 to 1, with higher values indicating greater structural similarity.

LPIPS measures perceptual similarity using deep feature representations extracted from a pretrained neural network. It is computed as:

$$\mathrm{LPIPS}\left(\hat{I}, I\right) = \sum_{l}\frac{1}{H_l W_l}\sum_{h,w}\left\|\omega_l \odot \left(\phi_l(\hat{I})_{h,w} - \phi_l(I)_{h,w}\right)\right\|_2^2 \tag{14}$$

where $\phi_l(\cdot)$ denotes the feature map at layer $l$, $\omega_l$ is a learned channel-wise weight, and $H_l$, $W_l$ are the spatial dimensions of the feature map. Lower LPIPS values indicate higher perceptual similarity.

By jointly employing PSNR, SSIM, and LPIPS, the evaluation captures complementary aspects of rendering quality, enabling a comprehensive assessment of both low-level accuracy and high-level perceptual realism in the reconstructed RGB images.

*4.2.2 Gaussian-center geometric consistency*

The geometric consistency of the Gaussian-center representation was evaluated relative to MVS-derived reference geometry. The point-wise depth map aggregated from Gaussian centers was compared with the reference depth map obtained by projecting the MVS dense point cloud into the corresponding camera views. The mean absolute error (MAE) is adopted as the primary metric for evaluating Gaussian-center depth consistency:

$$MAE_{Depth} = \frac{1}{|\Omega_D|} \sum_{x \in \Omega_D} \left| \widehat{D}(x) - D^{ref}(x) \right| \tag{15}$$

Since the depth supervision during training is formulated as an $L_1$ loss over the same valid region, MAE provides a consistent and direct measure of geometric accuracy for evaluating the alignment between Gaussian centers and reference surface geometry. This metric evaluates consistency with the MVS reference geometry rather than absolute physical accuracy. Physical measurement accuracy is examined separately through the HKUPB case study in Section 6.2.

*4.2.3 Semantic Gaussian field*

The semantic accuracy is evaluated by comparing the rendered semantic map $\widehat{M}$ with the ground-truth segmentation map $M$. The rendered semantic map is obtained by applying an argmax operation over the predicted semantic features at each pixel. Since the semantic field is learned in 3D and rendered into 2D, the evaluation reflects both semantic prediction accuracy and multi-view consistency. The primary metric is the mean Intersection over Union (mIoU), defined as:

$$\mathrm{mIoU} = \frac{1}{N_c}\sum_{c=1}^{N_c}\frac{\left|\widehat{M}_c \cap M_c\right|}{\left|\widehat{M}_c \cup M_c\right|} \tag{16}$$

where $N_c$ denotes the number of semantic classes, and $\widehat{M}_c$ and $M_c$ represent the predicted and ground-truth pixel sets of class $c$, respectively.

In addition, pixel accuracy (Acc) is reported as a complementary metric. While mIoU measures class-level overlap and is more sensitive to small or imbalanced categories, Acc provides an overall assessment of pixel-wise semantic correctness:

$$\mathrm{Acc} = \frac{\sum_x \mathbf{1}\left(\widehat{M}(x) = M(x)\right)}{N} \tag{17}$$

where $N$ denotes the total number of evaluated pixels, and $1(\cdot)$ is an indicator function.

4.3 Experimental setup

All experiments were conducted using Python 3.11 and PyTorch 2.8.0 on a workstation equipped with an NVIDIA GeForce RTX 5090 GPU with 32 GB of VRAM and CUDA 12.8. The proposed framework was optimized in two stages. In the first stage, RGB appearance was optimized within the mask-defined target region, and point-wise depth supervision was activated only after the photometric reconstruction had stabilized. In the second stage, the optimized geometry and appearance parameters were fixed, and only the per-Gaussian semantic features were learned from foreground segmentation labels.

The RGB reconstruction stage was trained for 40,000 iterations. The learning rates were fixed as $2\times10^{-5}$ for position, 0.01 for opacity, 0.006 for scaling, and 0.001 for rotation. Gaussian densification started from early iterations and stopped in the middle stage, with periodic opacity reset for pruning. For point-wise depth supervision, the maximum number of retained candidates $K$ was set to 3, the front-to-back decay coefficient $\tau$ to 0.95, the depth-consistency threshold $\delta$ to 1 in the common camera-space z-depth units of the Gaussian centers and MVS references, and the depth-loss weight $\lambda_d$ to 1. Gradients from the depth loss were applied only to Gaussian-center positions, while opacity, scaling, and rotation were not updated by this loss. In the second stage, the semantic field used an eight-dimensional feature vector and was optimized for 10,000 iterations.

# 5. Results and Discussion

## 5.1 Target-region RGB reconstruction

This section examined whether mask-based region supervision improves target-building reconstruction and whether the subsequent introduction of depth supervision preserves the photorealistic appearance of 3DGS. The evaluation therefore considered both target-region reconstruction quality and the appearance-geometry trade-off introduced by depth constraints. Table 2 reports the quantitative results of RGB reconstruction on the two datasets. For clarity, PCR denotes point cloud rendering, FG denotes foreground, BG denotes background, RS denotes region supervision, and DS denotes depth supervision. Overall, point cloud rendering produced substantially lower reconstruction accuracy than 3DGS methods on both datasets, indicating the limitation of direct point-based rendering in preserving continuous appearance and fine façade textures. In contrast, 3DGS achieved substantially higher PSNR and SSIM values, demonstrating its strong capability in photorealistic view synthesis.

Table 2. Quantitative comparison of RGB reconstruction quality.

| Method | QuakeCity | | | HKUPB | | |
|---|---|---|---|---|---|---|
| | PSNR ↑ | SSIM ↑ | LPIPS ↓ | PSNR ↑ | SSIM ↑ | LPIPS ↓ |
| PCR | 7.3979 | 0.6012 | 0.4109 | 15.6801 | 0.6470 | 0.3655 |
| 3DGS (FG & BG) | 29.7080 | 0.8161 | 0.3329 | 24.7176 | 0.7324 | 0.3354 |
| 3DGS (FG) | 33.9192 | 0.8568 | 0.1469 | 24.8151 | 0.7492 | 0.2723 |
| Ours w/ RS | 34.6252 | 0.8805 | 0.1170 | 26.2416 | 0.7646 | 0.2581 |
| Ours w/ RS & DS | 33.5551 | 0.8661 | 0.1273 | 25.0000 | 0.7500 | 0.2680 |

Note: PCR = point cloud rendering; FG = foreground; BG = background; RS = region supervision; DS = depth supervision; w/ denotes with.

A comparison between 3DGS (FG & BG) and 3DGS (FG) further reveals the adverse influence of background regions on RGB quality evaluation. Standard 3DGS optimizes all visible pixels in the input images, including both the target building and non-target background regions. However, background areas are often farther from the cameras and observed from few effective viewpoints, making them more difficult to reconstruct with stable visual quality. Consequently, including background regions in the evaluation may reduce the overall RGB metrics and underestimate the reconstruction quality of the target building itself. This effect is particularly evident on the QuakeCity dataset, where 3DGS (FG) outperforms 3DGS (FG & BG). On the HKUPB dataset, the improvement is relatively moderate, mainly because the smartphone-captured views are largely centered on the target building façade and contain fewer background regions. Therefore, the FG-only evaluation provides a more targeted assessment of building reconstruction performance.

The proposed model with region supervision, denoted as ours w/ RS, achieved the best overall RGB reconstruction performance on both datasets. On QuakeCity, it obtained the highest PSNR of 34.6252, SSIM of 0.8805, and the lowest LPIPS of 0.1170. On HKUPB, it also achieved the best results, with PSNR, SSIM, and LPIPS values of 26.2416, 0.7646, and 0.2581, respectively. These improvements demonstrate that region supervision enables the optimization process to focus more effectively on target regions while reducing the influence of irrelevant background areas. After further incorporating depth supervision, ours w/ RS & DS maintained competitive RGB reconstruction quality, although a slight decrease in RGB evaluation metrics can be observed compared with the RS-only model. This reduction is expected because depth supervision introduces an additional geometric constraint, encouraging Gaussian centers to move toward physically meaningful surface locations rather than being optimized solely for appearance fitting. Nevertheless, the RGB performance remains comparable to or better than the baseline 3DGS variants, indicating that the proposed framework can improve geometric reliability while largely preserving photorealistic reconstruction quality.

Qualitative comparisons are presented in Fig. 4. The input images show the reference appearance of the target buildings in the two datasets, while the rendered results compare PCR, standard 3DGS, and the proposed method under both original and full-scene views. PCR roughly preserves the geometry of the target building but shows limited capability in producing continuous and photorealistic textures. PCR mainly renders the point cloud of the target building, while many distant background regions are absent or incomplete due to insufficient point-cloud reconstruction. By contrast, standard 3DGS attempts to reconstruct both the target building and surrounding background from image supervision. Although this leads to more complete scene rendering, distant and sparsely observed background regions often appear blurred, noisy, or visually unstable.

On the QuakeCity dataset, the proposed method produces clear façade structures, component boundaries, and visually consistent textures. In the full-view comparison, irrelevant background regions are largely suppressed, allowing the reconstructed representation to remain concentrated on the target building. Similar behavior is observed on the real-world HKUPB dataset, where most background regions and foreground occlusions caused by adjacent building columns are removed while the principal façade appearance is retained. This result is not equivalent to simple image cropping, which only changes the spatial extent of the input image. Mask-based region supervision instead excludes non-target pixels during optimization, including occluding regions located within the image-space extent of the target building.

Nevertheless, residual local artifacts remain in the HKUPB results, particularly near the roof boundaries, where slight white highlights and Gaussian blurring can be observed. These artifacts can be attributed to two main factors. First, the handheld images were captured primarily from ground-level viewpoints, providing limited observations of the roof and upper façade boundaries. Second, unlike the precisely rendered labels available for QuakeCity, the HKUPB masks generated using SAM 3 may contain pixel-level deviations along complex object boundaries. Such annotation uncertainty can weaken region supervision near narrow edges and partially occluded regions. Although the synthetic QuakeCity results exhibit cleaner target boundaries, the HKUPB results indicate that the proposed strategy remains applicable under more challenging real-world acquisition conditions.

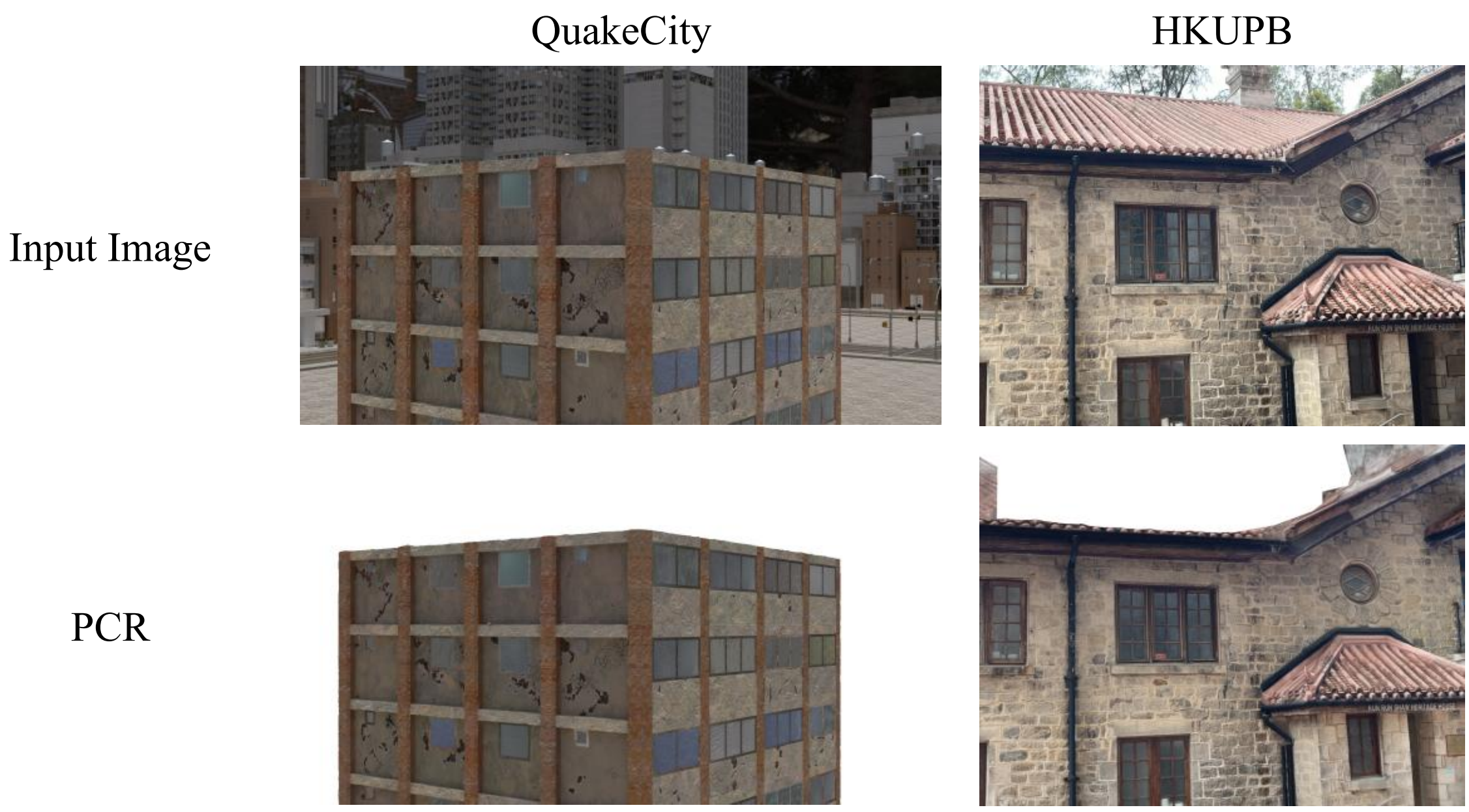

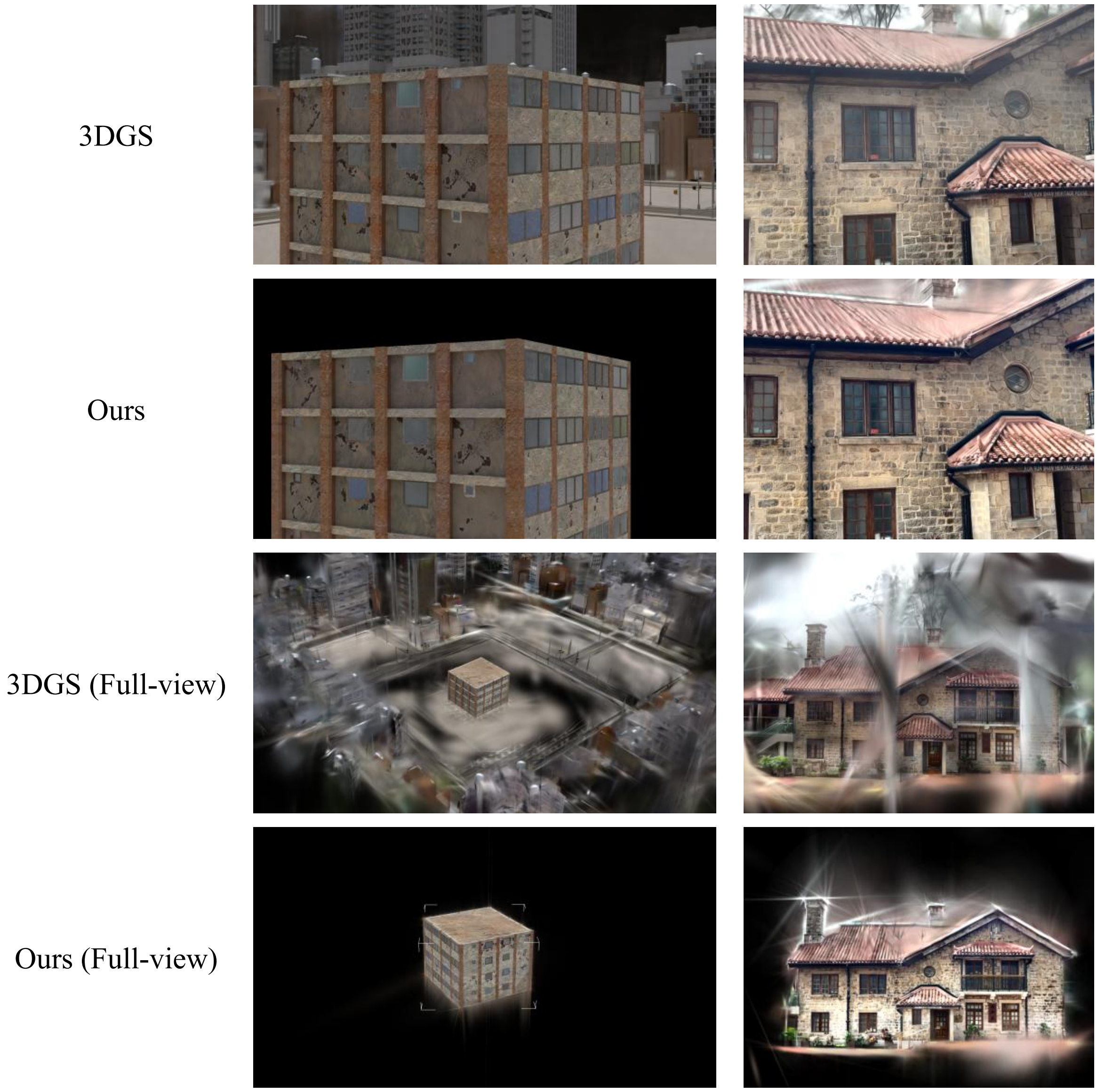

Fig. 4. Comparison of RGB reconstruction quality.

Overall, region supervision improves reconstruction within inspection-relevant building regions while reducing the influence of poorly observed backgrounds and foreground occlusions. The additional depth constraint introduces a limited reduction in RGB metrics but largely preserves the visual fidelity required for engineering inspection. These findings establish the appearance-level basis for the geometric and semantic evaluations that follow.

## 5.2 Gaussian-center geometric consistency

This section examined whether point-wise depth supervision improves the alignment of Gaussian centers with MVS-derived reference surfaces without substantially disrupting RGB optimization. As shown in Fig. 5, depth supervision was activated only after the RGB reconstruction had reached a relatively stable stage. The RGB loss remained generally stable after activation on both datasets, indicating that the additional geometric constraint did not substantially disrupt appearance optimization. Meanwhile, the depth loss decreased after depth supervision was introduced. The decrease was particularly pronounced for HKUPB, where uncontrolled smartphone acquisition, incomplete viewpoint coverage, and façade occlusions produced a larger initial discrepancy between Gaussian-center depths and the MVS-derived references. These loss curves demonstrate that the depth-supervision objective converged as intended. However, because the depth loss is calculated against the same MVS-derived reference used for supervision, it characterizes optimization behavior rather than independent physical measurement accuracy, which is examined separately in Section 6.2.

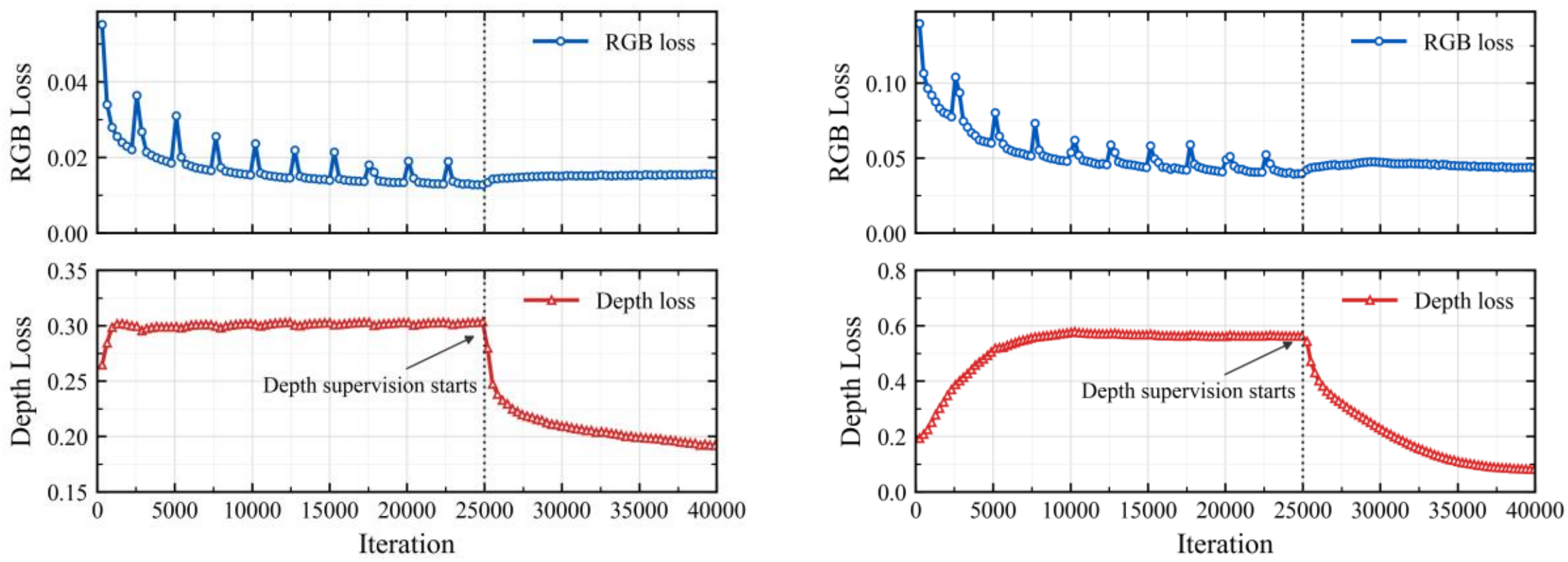

(a) QuakeCity (b) HKUPB

Fig. 5. Training losses of ours w/ RS & DS.

Table 3 quantifies Gaussian-center depth consistency against the MVS-derived reference geometry using MAE. Depth MAE was computed between the point-wise z-depth aggregated from Gaussian centers and the MVS-derived reference z-depth in the shared reconstruction coordinate system. The model incorporating both region and depth supervision achieved the lowest MAE on both datasets. On QuakeCity, the MAE decreased from 0.4575 for standard 3DGS to 0.2921 for Ours w/ RS and was further reduced to 0.1918 for Ours w/ RS & DS, corresponding to a reduction of approximately 58.1% relative to standard 3DGS. On HKUPB, the MAE decreased from 0.6159 for standard 3DGS and 0.5943 for Ours w/ RS to 0.0816 for Ours w/ RS & DS, corresponding to a reduction of approximately 86.8% relative to standard 3DGS. These results demonstrate improved consistency between Gaussian-center geometry and the MVS-derived reference surfaces.

Table 3. Quantitative comparison of Gaussian-center depth consistency.

| Method | Depth MAE (scene units) ↓ | |
|---|---|---|
| | QuakeCity | HKUPB |
| 3DGS | 0.4575 | 0.6159 |
| Ours w/ RS | 0.2921 | 0.5943 |
| Ours w/ RS & DS | 0.1918 | 0.0816 |

Fig. 6 presents the qualitative comparison results, where the blue dots denote the projected Gaussian centers. On the QuakeCity dataset, the comparison between Fig. 6(a) and Fig. 6(b) shows that depth supervision improves the spatial alignment of Gaussian centers along the two dominant planar directions of the building façades. In the RS-only model, the Gaussian centers are relatively scattered around the building surfaces, with some primitives deviating from the dominant planar geometry. After incorporating depth supervision, the Gaussian centers adhere more closely to the façade planes and form a more coherent surface distribution. Similar but more pronounced improvements can be observed on the HKUPB dataset, as shown in Fig. 6(c) and Fig. 6(d). Owing to uncontrolled smartphone capture, incomplete view coverage, foreground occlusions, and complex façade geometry, the RS-only model contains a considerable number of floating and geometrically unstable primitives. With depth supervision, these scattered points are substantially reduced, and the Gaussian centers become more densely concentrated around the visible building surfaces. This effect is particularly evident in Fig. 6(d), where the optimized primitives exhibit clearer alignment with the building façades and their primary substructures. Overall, the visual comparison shows that the proposed depth supervision reduces floating primitives and improves the alignment of Gaussian centers with the MVS-derived reference surfaces, providing a more coherent geometric basis for subsequent measurement.

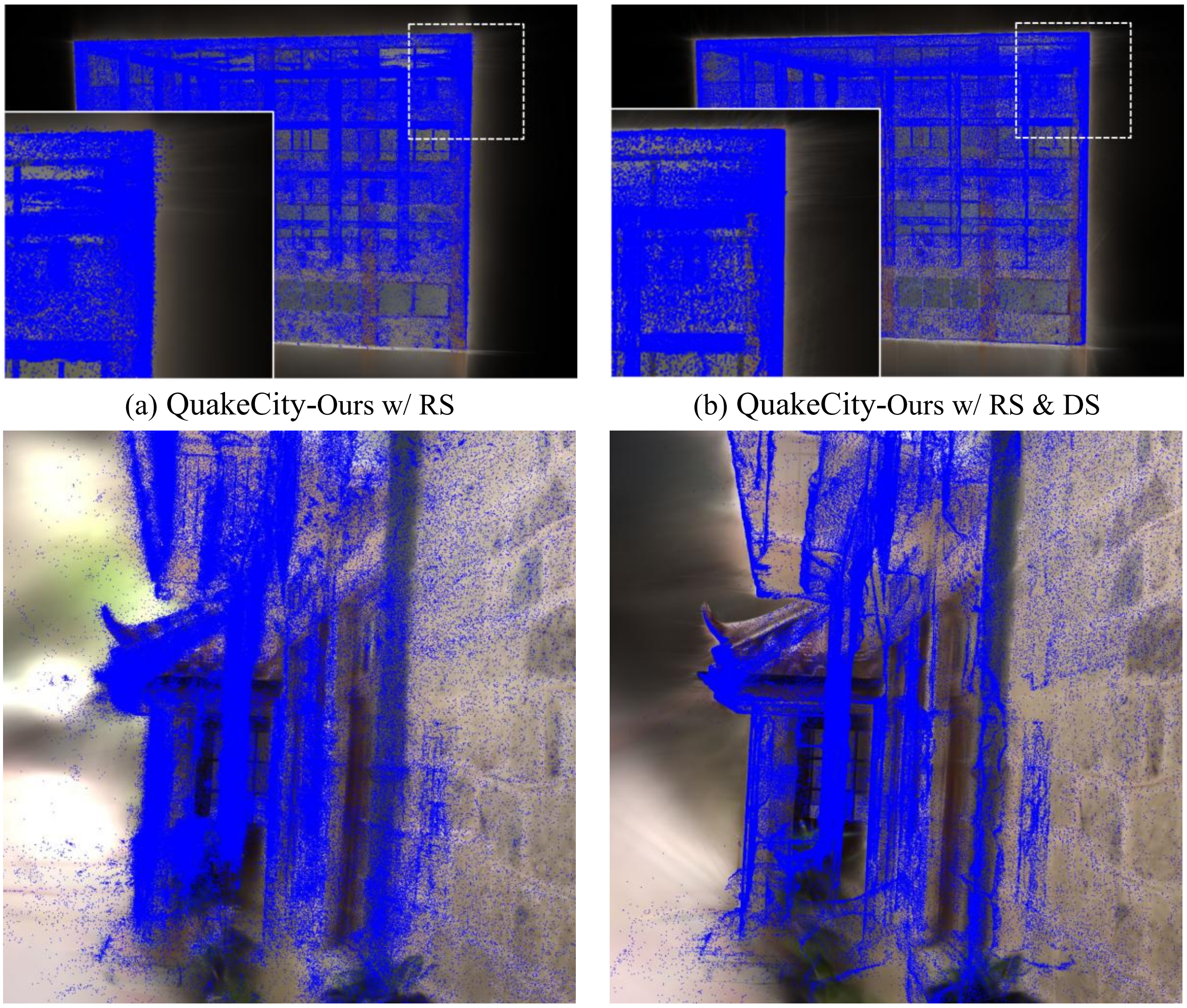

(a) QuakeCity-Ours w/ RS (b) QuakeCity-Ours w/ RS & DS

(c) HKUPB-Ours w/ RS (d) HKUPB-Ours w/ RS & DS

Fig. 6. Comparison of depth supervision results.

Taken together, the quantitative and qualitative results show that region supervision first removes irrelevant optimization domains, while depth supervision subsequently concentrates Gaussian centers around the reference building surfaces. This establishes a more coherent geometric basis for point selection and surface measurement. The extent to which this improved consistency translates into physical measurement accuracy is evaluated separately in Section 6.2.

### 5.3 Semantic rendering and component-level interpretation

This experiment examined whether semantic information attached to individual Gaussian primitives can be rendered consistently across views and used to distinguish building components and defect categories. To evaluate the quality of the constructed semantic Gaussian field, both quantitative metrics and visual comparisons are reported for the QuakeCity and HKUPB datasets. Table 4 presents the class-wise IoU, mIoU, and Acc of the rendered semantic images (Pre) against the corresponding ground-truth labels (GT). Overall, the proposed semantic Gaussian field achieves satisfactory semantic rendering performance on both datasets, with mIoU values of 0.8529 and 0.8642 for QuakeCity and HKUPB, respectively. The corresponding Acc values reach 0.9881 and 0.9815, indicating that most pixels are correctly assigned to their semantic categories after being rendered from the learned 3D Gaussian representation.

For the QuakeCity dataset, most categories achieve high IoU values, including background, wall, beam, column, window frame, window, and spalling. As shown in Fig. 7, the predicted semantic renderings are highly consistent with the ground-truth labels in terms of both global façade layout and local component distribution. Major building elements, such as walls, beams, columns, and windows, are accurately rendered, and their spatial relationships are well preserved in the semantic field. However, the IoU of cracking is relatively low. This is mainly because cracking regions are thin, sparse, and locally distributed, making them sensitive to boundary deviations and rendering variations. Moreover, cracking pixels occupy only a small proportion of the labeled region and therefore provide limited supervision during semantic-field optimization, resulting in substantially lower IoU for this category.

Table 4. Quantitative comparison of semantic rendering results.

| QuakeCity | | HKUPB | |
|---|---|---|---|
| Classes | IoU ↑ | Classes | IoU ↑ |
| Background | 0.9945 | Background | 0.8800 |
| Wall | 0.9708 | Wall | 0.8824 |
| Beam | 0.9641 | Column | 0.9702 |
| Column | 0.9646 | Nameplate | 0.8230 |
| Window frame | 0.9081 | Nameplate2 | 0.7480 |
| Window | 0.9782 | Round window | 0.8323 |
| Cracking | 0.1988 | Window | 0.9133 |
| Spalling | 0.8440 | - | - |
| mIoU ↑ | 0.8529 | mIoU ↑ | 0.8642 |
| Acc ↑ | 0.9881 | Acc ↑ | 0.9815 |

For the HKUPB dataset, the semantic field also demonstrates strong performance under more challenging real-world conditions. The dataset contains complex stone textures, irregular façade surfaces, roof structures, vegetation occlusions, non-uniform illumination, and smartphone-based viewpoints, which make semantic reconstruction more difficult than the synthetic QuakeCity dataset. Despite these challenges, the proposed method achieves an mIoU of 0.8642 and accurately reconstructs major semantic categories such as background, wall, column, and window. The visual results in Fig. 7 further confirm that the rendered semantic images are generally consistent with the ground-truth annotations and that semantic regions remain spatially coherent when overlaid onto the RGB renderings.

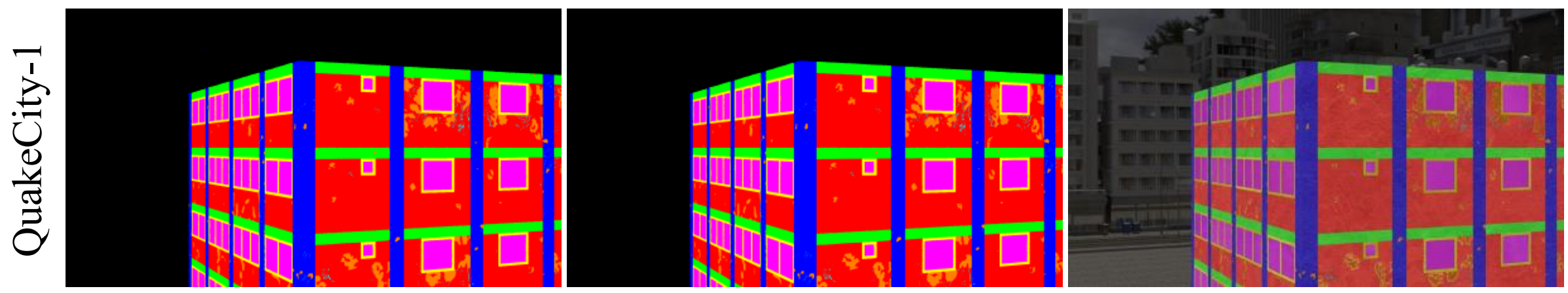

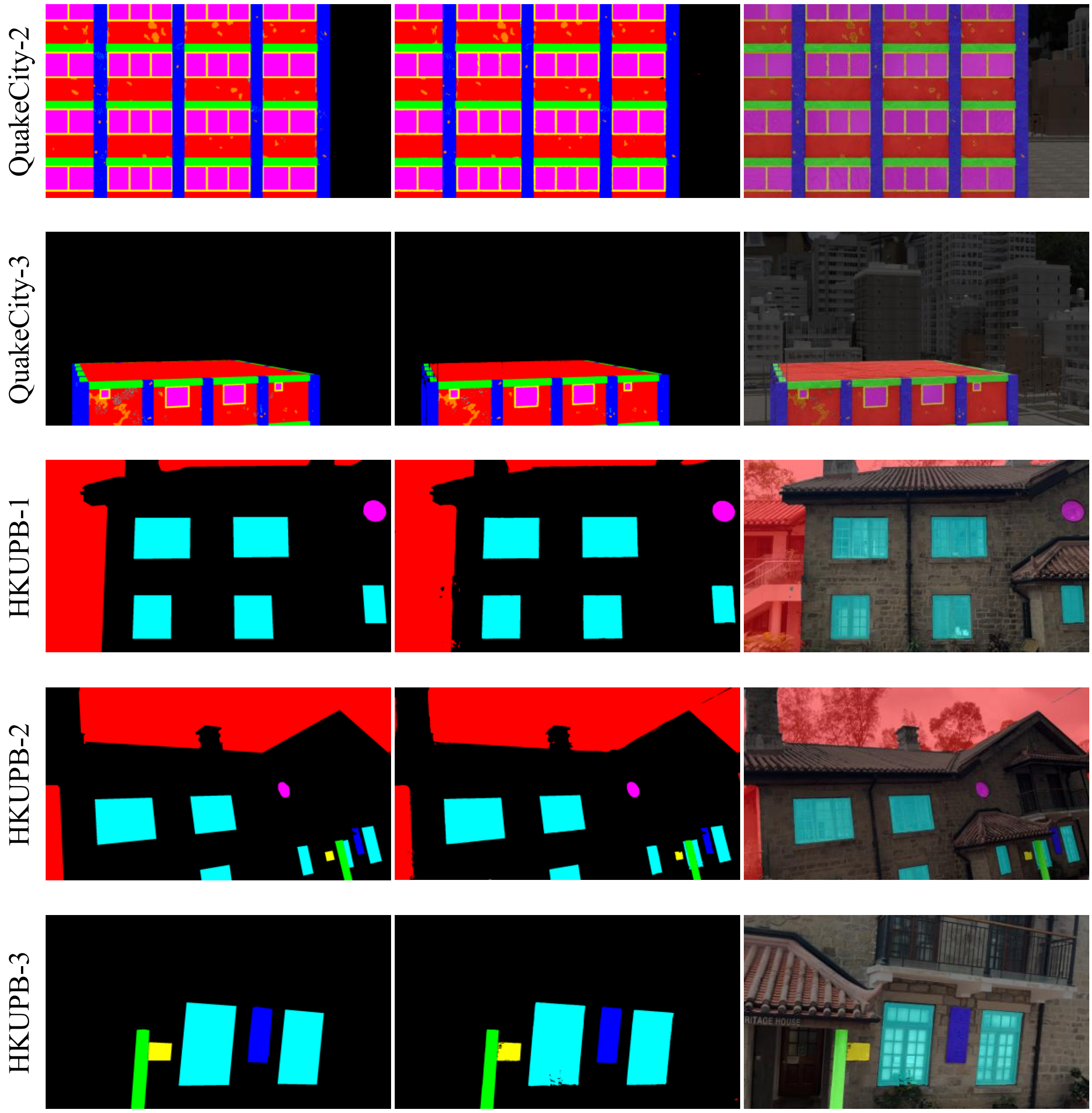

(a) GT (b) Pre (c) Overlay

Fig. 7. Comparison of semantic rendering results.

Nevertheless, the two nameplate categories and the round-window category exhibit lower IoU values than the major façade components. These categories occupy smaller image regions, are visible from fewer viewpoints, and contain boundaries that can be difficult to distinguish from the surrounding stone façade. In addition, the HKUPB annotations were generated using a SAM 3 annotation tool, which may introduce pixel-level deviations along complex object edges. Such deviations have a proportionally larger influence on the IoU of small or narrow categories. Correspondingly, minor boundary inconsistencies and fragmented predictions can be observed near roof edges, occluded areas, and small façade details in Fig. 7.

Overall, the quantitative and qualitative results show that semantic information can be attached to Gaussian primitives and consistently rendered into image space. The semantic field achieves stable performance under the controlled QuakeCity setting and retains useful component-level representation under the more challenging HKUPB conditions. By associating semantic classes with the same Gaussian primitives used for visualization and measurement, the proposed framework extends 3DGS from photorealistic reconstruction toward a structured scene representation suitable for semantic inspection and category-guided engineering operations.

# 6. Engineering applications of AEC-SplatTwin

Following the representation-level evaluation in Section 5, this section demonstrates how the proposed visual, geometric, and semantic capabilities can be combined within AEC-SplatTwin to support engineering inspection tasks. The demonstrations follow a task-oriented workflow comprising multimodal scene inspection, quantitative measurement, and MLLM-assisted information access.

## 6.1 Real-time multimodal inspection

To support interactive inspection and engineering-oriented analysis, AEC-SplatTwin provides a real-time multimodal visualization interface for optimized semantic 3D Gaussian representations. As illustrated in Fig. 8, the system supports multiple visualization modes on both the QuakeCity and HKUPB datasets, including RGB rendering, semantic rendering, semantic overlay, comparative visualization, point-based Gaussian inspection, point-level semantic category visualization, and point-level semantic instance analysis. These modes enable users to examine the reconstructed scene from complementary perspectives, covering photorealistic appearance, semantic structure, Gaussian primitive distribution, and component-level organization.

QuakeCity HKUPB

(a) RGB rendering

(b) Semantic rendering

(c) Semantic overlay

(d) Comparison view

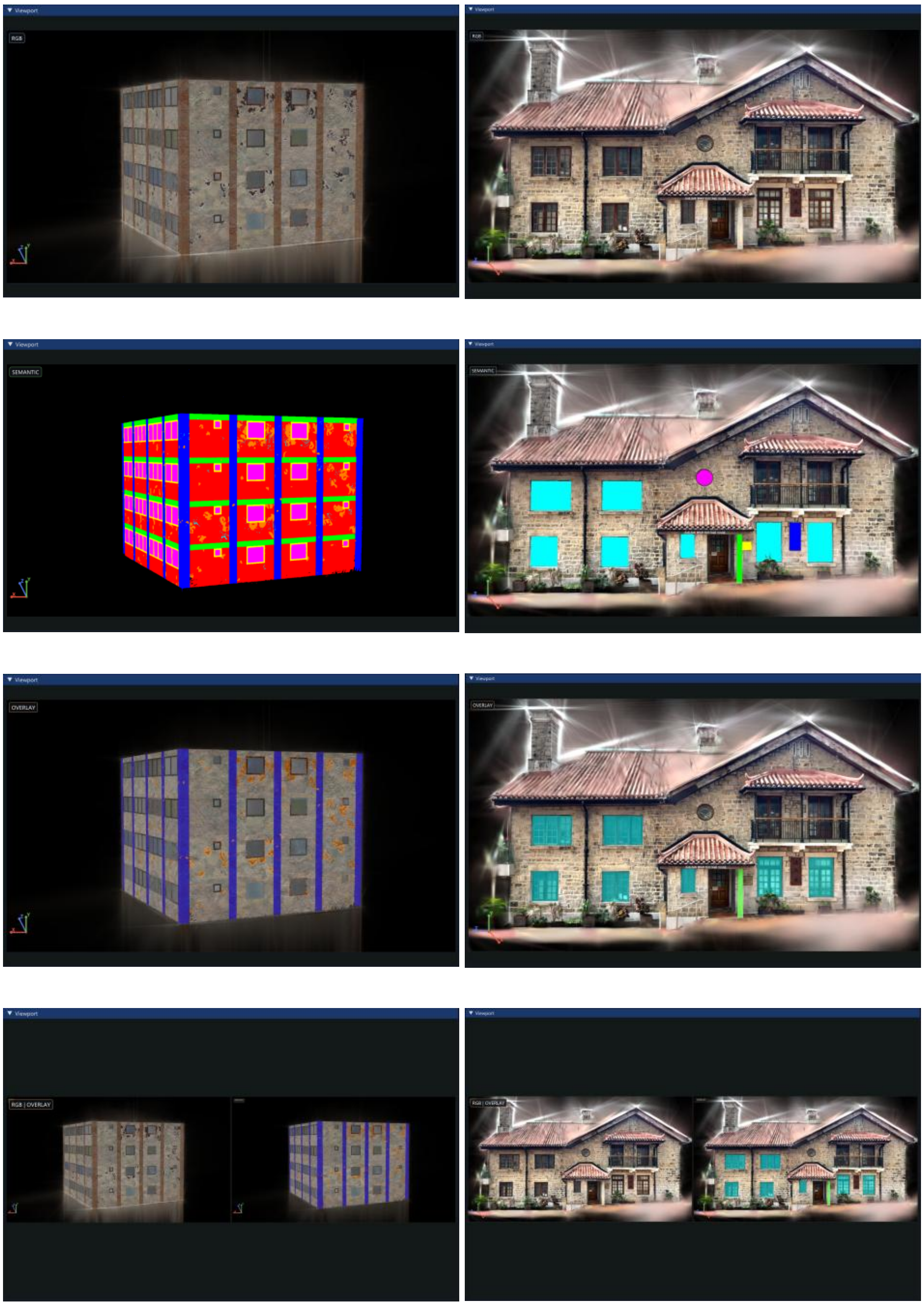

(e) Gaussian points

(f) Point-level semantic class

(g) Point-level semantic instances

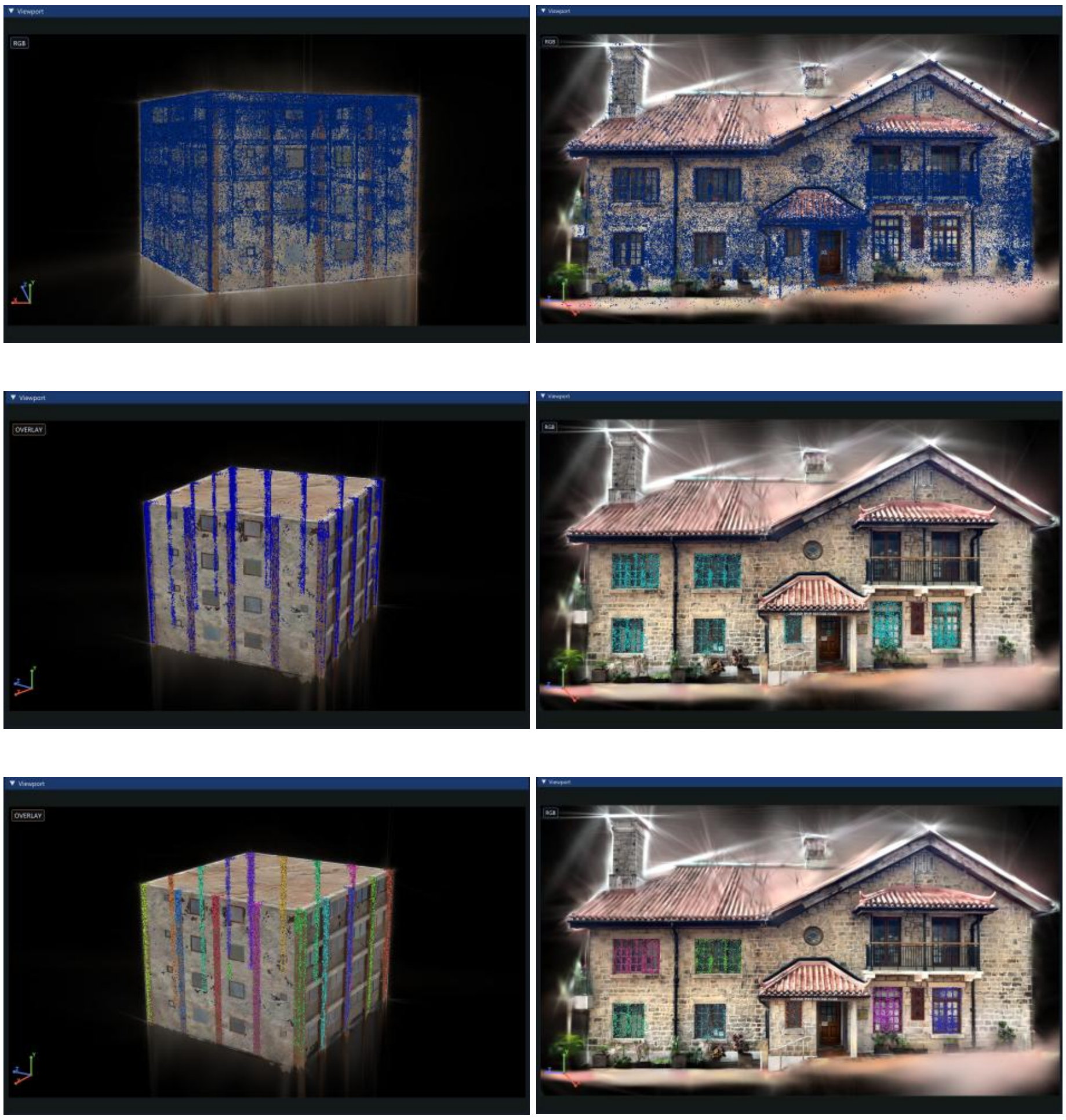

Fig. 8. Real-time multimodal visualization of AEC-SplatTwin across RGB, semantic, overlay, comparison, point, and instance-level modes.

In the RGB rendering mode, the optimized 3DGS representation is rendered in real time, enabling users to intuitively inspect the photorealistic reconstruction quality of the target building. At a rendering resolution of 2560 × 1440 pixels, AEC-SplatTwin maintained a stable frame rate of approximately 60 FPS, enabling real-time rendering and continuous visual inspection from both predefined camera trajectories and freely controlled viewpoints. In the semantic rendering mode, the learned semantic Gaussian field is visualized by rendering category-level predictions directly onto the current view. For the QuakeCity dataset, semantic rendering clearly distinguishes façade components and damage categories, whereas for the HKUPB dataset, major architectural elements such as window, nameplate, and column can be interactively highlighted. The semantic overlay mode further combines semantic predictions with RGB rendering, allowing users to jointly assess semantic consistency and visual reconstruction quality. This is particularly useful for verifying whether the predicted semantic regions are spatially aligned with the corresponding components. In addition, the comparison view provides side-by-side visualization of different rendering results, such as RGB reconstruction and semantic overlay. This function enables users to directly compare appearance and semantic representations from the same viewpoint.

Beyond image-level visualization, AEC-SplatTwin also provides point-based inspection of the underlying Gaussian primitives. In the Gaussian point view, Gaussian centers are projected into the viewport, allowing users to examine the discrete geometric primitives that constitute the reconstructed scene. Based on the semantic field, the point-level semantic category mode further assigns category information to each Gaussian center, enabling users to inspect the spatial distribution of semantic categories at the primitive level. This establishes a connection between image-level semantic rendering and 3D point-level scene understanding. Furthermore, the point-level semantic instance mode applies clustering algorithms to Gaussian centers, grouping them into distinct semantic instances. As shown in Fig. 8(g), different instances belonging to the same semantic category can be visually distinguished, which facilitates more detailed analysis of repeated architectural components, such as columns and windows. This capability extends semantic 3DGS from category-level visualization to instance-aware inspection.

These visualization modes form an engineering inspection sequence rather than a collection of independent displays: users can first verify the reconstructed appearance, then locate target components or defects through semantic overlays, and finally inspect the corresponding Gaussian centers before performing geometric measurement.

6.2 Engineering measurement validation

Beyond real-time visualization and semantic inspection, AEC-SplatTwin further supports interactive measurement for engineering-oriented analysis. As shown in Fig. 9, the system provides two complementary measurement modes: point-based distance measurement and pixel-based area measurement. In the point-based mode, users can directly select Gaussian centers in the rendered scene and calculate the spatial distance between two selected points. In contrast, the pixel-based mode estimates real-world areas from the current rendered view. By combining the rendered depth map and camera parameters, the spatial scale represented by image pixels can be estimated, allowing a selected region or semantic category to be converted into a physical area.

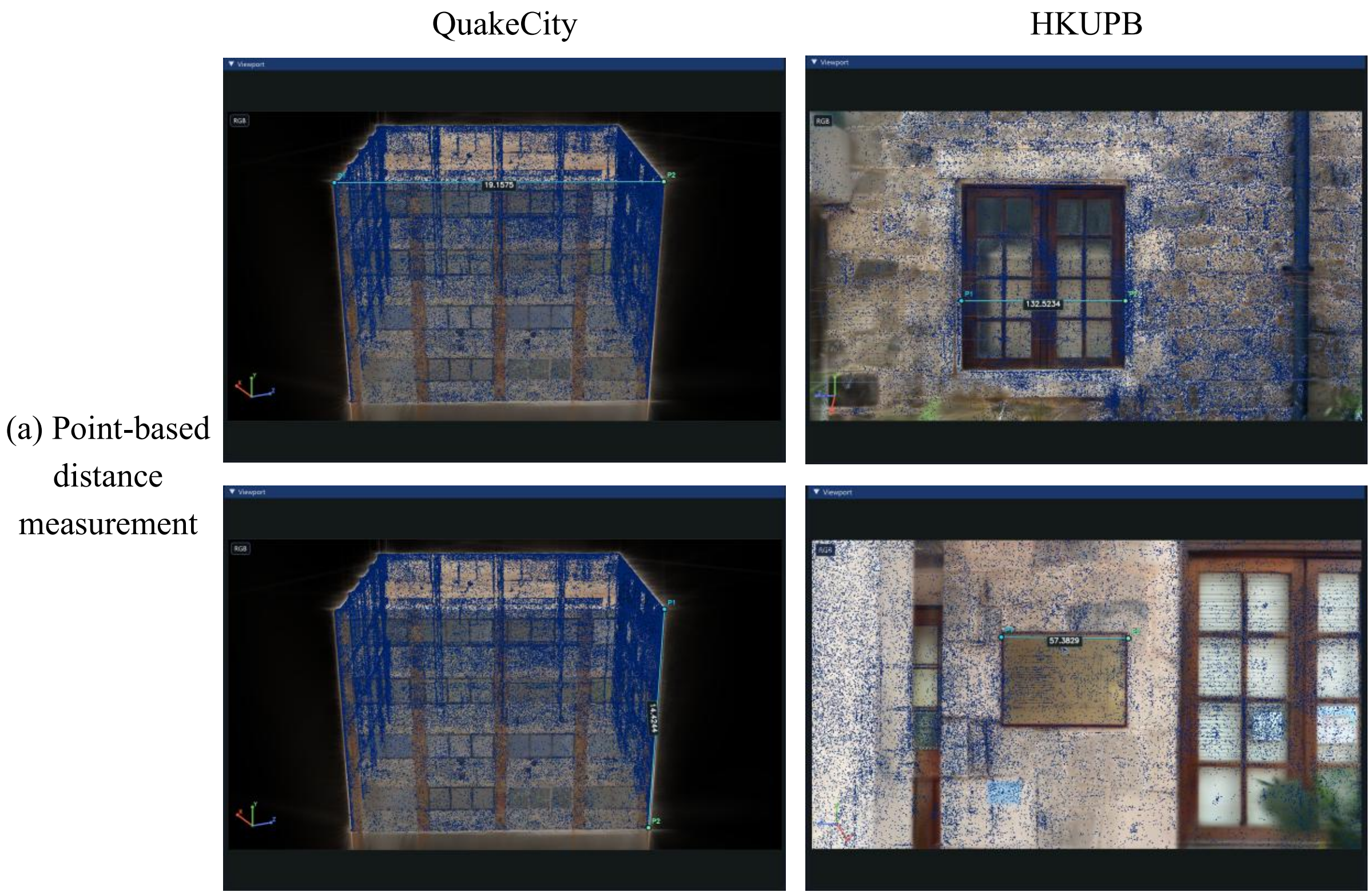

(b) Pixel-based area measurement

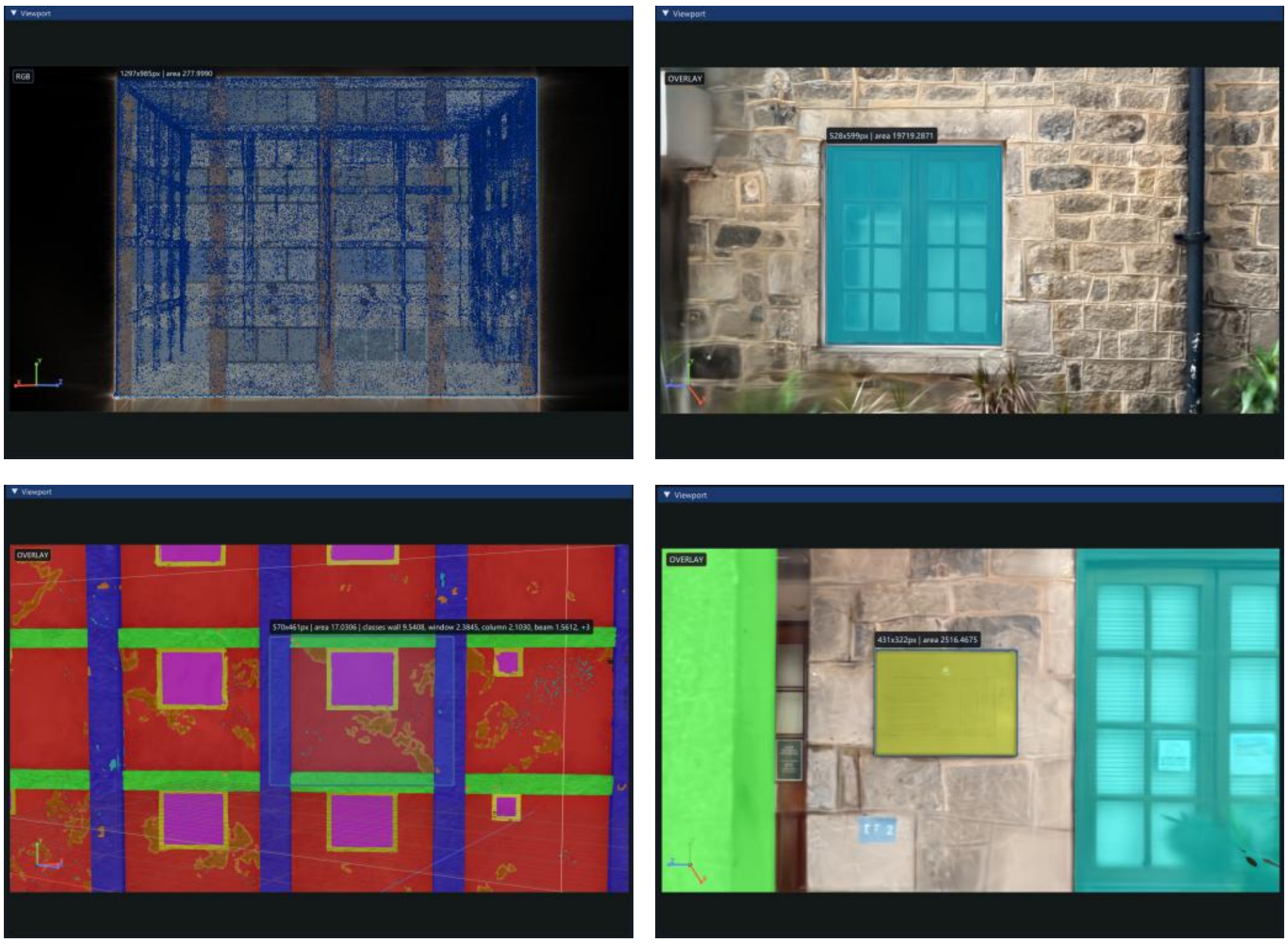

Fig. 9. Interactive measurement of AEC-SplatTwin across point-based distance and pixel-based area modes.

The quantitative measurement results are summarized in Table 5. It should be noted that different scale settings were adopted for the two datasets. Since the QuakeCity dataset does not provide real-world physical dimensions, it was mainly used to evaluate the internal consistency between point-based distance measurement and pixel-based area measurement for the façade shown in Fig. 9. For HKUPB, the measured height and width of a separate window excluded from Table 5 were used for scale calibration. The calibrated scale was then applied unchanged to seven independent targets whose estimated dimensions and areas were compared with manual measurements. For point-based distance measurement, AEC-SplatTwin achieves high consistency with reference values. Across the evaluated targets, the average relative errors are 0.33% for height and 1.07% for width, indicating that the depth-supervised Gaussian representation provides sufficient geometric reliability for interactive distance measurement. For pixel-based area measurement, the proposed system also demonstrates satisfactory accuracy, with an average relative error of 0.60%. This result shows that the pixel-based measurement mode can provide reliable surface-area estimation from the rendered view. Moreover, the consistently low errors across both the synthetic QuakeCity scene and the real-world HKUPB scene suggest that the proposed measurement strategy is applicable under different data conditions.

Table 5. Quantitative evaluation of interactive measurement.

| | | AEC-SplatTwin | | | Reference measurement | | | Relative Error | | |
|---|---|---|---|---|---|---|---|---|---|---|
| | | Point | | Pixel | | | | Point | | Pixel |
| | | H/cm | W/cm | Area/cm$^2$ | H/cm | W/cm | Area/cm$^2$ | H | W | Area |
| QuakeCity | Facade | 14.42 | 19.16 | 277.99 | - | - | 276.29* | - | - | 0.62% |
| | Window 1 | 149.38 | 132.52 | 19719.29 | 149 | 132.5 | 19742.5 | 0.26% | 0.02% | 0.12% |
| | Window 2 | 149.72 | 133.87 | 20006 | 150 | 135 | 20250 | 0.19% | 0.84% | 1.20% |
| | Window 3 | 199.17 | 125.46 | 25413.75 | 200 | 126.5 | 25300 | 0.42% | 0.82% | 0.45% |
| HKUPB | Window 4 | 200.15 | 127.04 | 25142.34 | 200 | 126.5 | 25300 | 0.08% | 0.43% | 0.62% |
| | Column | 232.2 | 21.5 | 5124.29 | 232 | 22 | 5104 | 0.09% | 2.27% | 0.40% |
| | Nameplate 1 | 139.02 | 52.61 | 7633.38 | 140 | 53.8 | 7532 | 0.70% | 2.21% | 1.35% |
| | Nameplate 2 | 43.23 | 57.38 | 2516.47 | 43.5 | 57.9 | 2518.65 | 0.62% | 0.90% | 0.09% |
| Average | | - | - | - | - | - | - | 0.33% | 1.07% | 0.60% |

Note: * denotes the reference area of the QuakeCity façade, which was calculated as the product of the H and W obtained from point-based measurement.

Overall, the results demonstrate how AEC-SplatTwin converts the optimized semantic 3DGS representation into an interactive measurement environment. The point-based distance mode enables direct geometric measurement based on optimized Gaussian centers, while the pixel-based area mode supports area estimation by combining depth information, camera parameters, and semantic results. The low relative errors reported in Table 5 indicate that the proposed system can support practical quantitative inspection tasks. More importantly, the coexistence of point-based and pixel-based measurement modes provides an internal consistency check, thereby improving the reliability and interpretability of interactive measurement in engineering applications.

## 6.3 MLLM-assisted interaction

To improve access to the visual, semantic, and geometric information maintained by AEC-SplatTwin, an MLLM-assisted interaction module is integrated into the system. Four representative functions are demonstrated in Fig. 10. First, natural-language instructions can be mapped to viewer operations, such as switching visualization modes, filtering semantic categories, and highlighting selected classes. Second, the current rendered view and semantic results can be supplied to the MLLM to generate a summary of visible components and their spatial distribution. Third, deterministic measurement outputs can be combined with the corresponding rendered view and semantic information to explain the measured object, selected region, and calculated result. The MLLM does not generate the geometric measurements; it interprets results produced by the measurement module. Finally, user-defined rules can be applied to the semantic and measurement outputs. As an illustrative example, the system reports a warning when the spalling-area ratio exceeds a predefined value of 5% within the selected region. This value is used only to demonstrate the rule-execution workflow and should not be interpreted as a generally validated engineering threshold.

Overall, the MLLM-assisted module provides a natural-language interface to the visualization, semantic inspection, and deterministic measurement functions of AEC-SplatTwin. It connects low-level Gaussian representations with task-oriented descriptions and user-defined inspection rules, reducing the effort required to access scene information. The module should be understood as a human-in-the-loop information-access and decision-support interface rather than an autonomous engineering assessment system. Engineering conclusions and risk classifications remain dependent on validated measurement procedures, project-specific rules, and professional review.

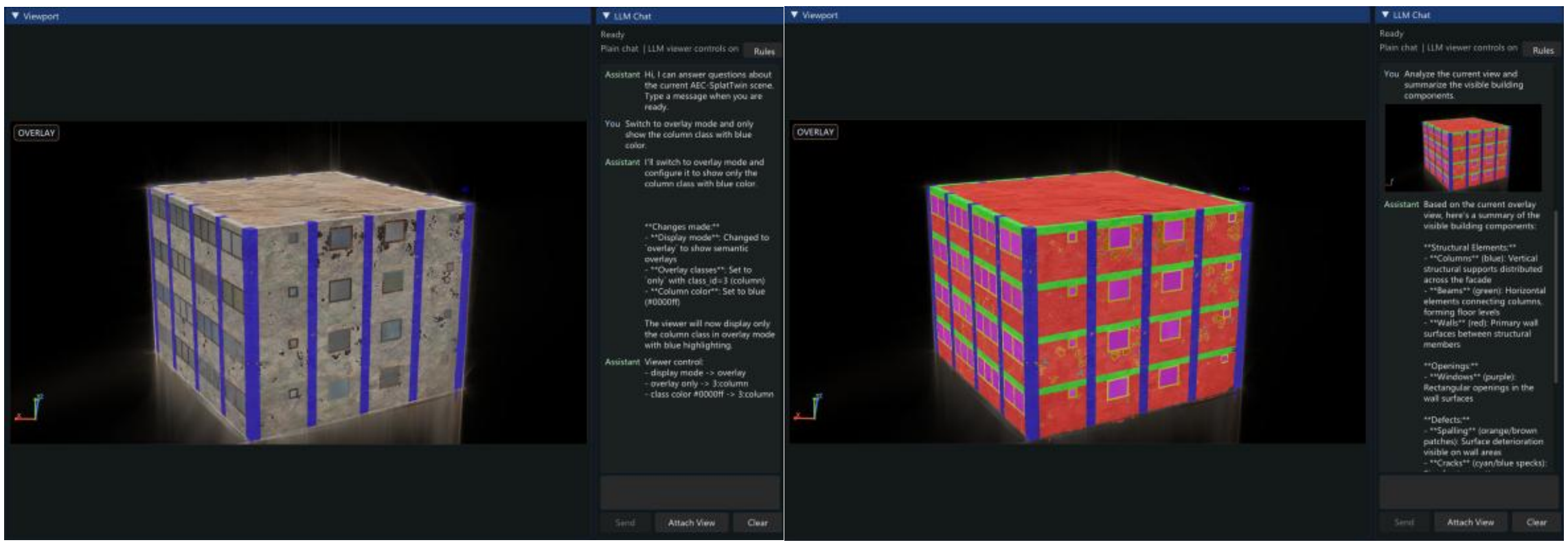

(a) Natural language control (b) Global scene understanding

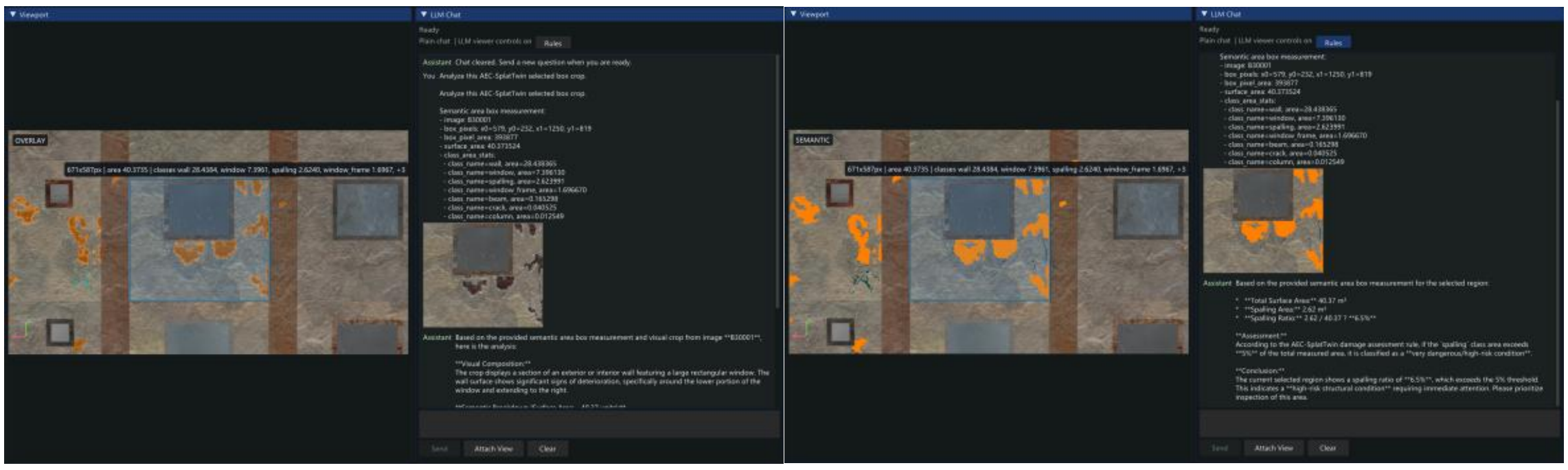

(c) Measurement result interpretation (d) Rule-based inspection

Fig. 10. MLLM-assisted interaction and operation in AEC-SplatTwin.

## 7. Limitations and future work

Despite the promising results, several limitations remain in the current framework. First, the spatial distribution of Gaussian primitives is not yet fully controllable during the optimization process. Although depth supervision improves the overall alignment between Gaussian centers and building surfaces, the final locations of Gaussian centers are determined by the training dynamics of 3DGS. As a result, some local regions may contain insufficient Gaussian centers, which can affect point-based measurement when users need to select exact points on specific structural boundaries. Future work may explore more controllable Gaussian initialization, densification, and refinement strategies to improve the robustness of point-level measurement.

Second, the reconstruction and semantic representation of thin, small, or highly localized structures remain challenging. For example, the cracking class usually occupies only a very small proportion of the image and may be underrepresented during both Gaussian optimization and semantic field construction. This can lead to incomplete reconstruction, blurred boundaries, or relatively weak semantic prediction for such fine-grained categories. Future studies could investigate local high-resolution reconstruction strategies, patch-based optimization, multi-scale semantic supervision, or damage-aware sampling mechanisms to enhance the representation of small and dense structural details.

Third, the current framework still relies on conventional photogrammetric reconstruction to obtain camera parameters and depth information. In particular, the depth supervision used in this study is derived from dense point cloud reconstruction, which may be affected by image quality, viewpoint coverage, texture conditions, and reconstruction completeness. This dependency may limit the applicability of the framework in scenes where reliable point clouds are difficult to obtain. In future work, recent foundation models for 3D reconstruction and depth estimation, such as VGGT and Depth Anything [56,57], could be incorporated to provide camera poses, depth priors, or geometry constraints. This may further simplify data preparation and improve the scalability of measurable semantic 3DGS for practical building inspection applications.

Overall, these limitations suggest that future development should focus on controllable Gaussian geometry, fine-grained structural representation, and more flexible sources of geometric supervision. Future work will further explore the integration of the proposed framework with robotic platforms for automated inspection and autonomous navigation, particularly in hazardous or hard-to-access scenarios, where measurable 3DGS representations can support scene reconstruction, path planning, and engineering-oriented measurement.

## 8. Conclusions

This study proposes a measurable semantic 3D Gaussian Splatting framework for interactive building inspection using multi-view inspection imagery. The framework integrates four connected capabilities. Mask-based region supervision defines a shared target domain for RGB and depth optimization, point-wise depth supervision constrains Gaussian-center depths using MVS-derived reference geometry, the semantic Gaussian field associates class information with individual Gaussian primitives, and AEC-SplatTwin operationalizes the resulting representation through real-time visualization, semantic inspection, geometric measurement, and MLLM-assisted interaction. Together, these developments extend 3DGS from a primarily photorealistic representation toward a target-focused, measurable, semantically structured, and interactive engineering representation for building inspection.

The framework was evaluated under two complementary inspection conditions represented by the synthetic UAV-based QuakeCity dataset and the real-world handheld HKUPB dataset. Region supervision improved target-region RGB reconstruction while suppressing irrelevant backgrounds and foreground occlusions. Point-wise depth supervision reduced the Gaussian-center depth MAE against the MVS-derived references from 0.4575 to 0.1918 on QuakeCity and from 0.6159 to 0.0816 on HKUPB, corresponding to reductions of 58.1% and 86.8%, respectively. The semantic Gaussian field achieved mIoU values of 0.8529 and 0.8642 on QuakeCity and HKUPB, respectively, supporting component- and defect-level semantic rendering. For the real-world HKUPB case, point-based height and width measurements achieved average relative errors of 0.33% and 1.07%, respectively, while pixel-based area measurement achieved an average relative error of 0.60%.

AEC-SplatTwin further demonstrates how visual appearance, semantic entities, geometric measurements, and natural-language information access can be maintained within a unified engineering workflow. Its MLLM-assisted module interprets deterministic system outputs and user-defined rules rather than directly generating measurement values or autonomous engineering decisions. Overall, the results establish a practical pathway for transforming image-based 3DGS reconstruction into an interactive engineering representation for building inspection applications. Future work should address controllable Gaussian distribution, improved representation of thin and small defects, reduced dependence on MVS-derived geometry, broader multi-building measurement validation, and integration with robotic inspection platforms in hazardous or difficult-to-access environments.

## CRediT authorship contribution statement

**Yi Zhang:** Writing – review & editing, Writing – original draft, Visualization, Validation, Methodology, Software, Data curation, Conceptualization. **Tian Tan:** Writing – review & editing, Software, Data curation, Conceptualization. **Ray Kai Leung Su:** Writing – review & editing, Project administration, Supervision.

## Statements and Declarations

### Ethical considerations

Not applicable.

### Consent to participate

Not applicable.

Consent for publication

Not applicable.

Declaration of conflicting interest

The authors declared no potential conflicts of interest with respect to the research, authorship, and/or publication of this article.

Funding statement

This research received no specific grant from any funding agency in the public, commercial, or not-for-profit sectors.

Data availability

The data supporting the findings of this study are available from the corresponding author upon reasonable request.